\documentclass{article}

\usepackage{multicol}
\usepackage[bookmarks=true]{hyperref}
\usepackage[english]{babel}
\usepackage{amsmath,amssymb,amsfonts}
\usepackage{algorithmic}
\usepackage{graphicx}
\usepackage{textcomp}
\usepackage{xcolor}
\usepackage{tabularx}
\usepackage{xspace}
\usepackage{url}
\usepackage{booktabs}
\usepackage{threeparttable}
\usepackage{array}
\usepackage{caption}
\usepackage{pifont}
\usepackage{bm}

\usepackage{enumitem}
\usepackage{gensymb}

\usepackage[final]{corl_2025} 

\definecolor{royalblue}{RGB}{65, 105, 225}
\definecolor{softgreen}{RGB}{85, 170, 85} 
\definecolor{softred}{RGB}{200, 50, 50}   

\hypersetup{
    colorlinks=true,
    linkcolor=orange,    
    citecolor=softgreen,    
    urlcolor=royalblue      
}

\newcommand{\system}{ToddlerBot\xspace}
\newcommand{\systems}{ToddlerBot's\xspace}
\newcommand{\tick}{\textcolor{softgreen}{\ding{51}}} 
\newcommand{\cross}{\textcolor{softred}{\ding{55}}}  

\title{\system: Open-Source ML-Compatible Humanoid Platform for Loco-Manipulation}

\author{
    \vspace{1mm}
    Haochen Shi$^{\textbf{*}}$ \quad
    Weizhuo Wang$^{\textbf{*}}$ \quad
    Shuran Song$^{\textbf{\dag}}$ \quad 
    C. Karen Liu$^{\textbf{\dag}}$ \\
    \vspace{1mm}
    Stanford University \\
    \vspace{1mm}
    $^{\textbf{*}}$Equal contribution \quad
    $^{\textbf{\dag}}$Equal advising \\
    \vspace{1mm}
    \texttt{\textcolor{magenta}{\url{https://toddlerbot.github.io}}}
    \vspace{-5mm}
}

\begin{document}
\maketitle


\begin{abstract}
Learning-based robotics research driven by data demands a new approach to robot hardware design—one that serves as both a platform for policy execution and a tool for embodied data collection. We introduce \system, a low-cost, open-source humanoid robot platform designed for robotics and AI research. \system enables seamless acquisition of high-quality simulation and real-world data. The plug-and-play zero-point calibration and transferable motor system identification ensure a high-fidelity digital twin and zero-shot sim-to-real policy transfer. A user-friendly teleoperation interface streamlines real-world data collection from human demonstrations. 
With its data collection ability and anthropomorphic design, \system is ideal for whole-body loco-manipulation research. Additionally, \systems compact size ($0.56~\mathrm{m},\space 3.4~\mathrm{kg}$) ensures safe operation in real-world environments. 
Reproducibility is achieved with entirely 3D-printed, open-source design and off-the-shelf components, keeping the total cost under $6,000~\mathrm{USD}$. This allows assembly and maintenance with basic technical expertise, as validated by successful independent replications of the system. We demonstrate \systems capabilities through arm span, payload, endurance tests, loco-manipulation tasks, and a collaborative long-horizon scenario where two robots tidy a toy session together. By advancing ML-compatibility, capability, and reproducibility, \system provides a robust and scalable platform for policy learning and execution in robotics research.

\end{abstract}

\keywords{Humanoid, Mechanisms \& Design, Robot Modeling \& Simulation} 

\begin{figure}[t]
  \centering
\includegraphics[width=\textwidth]{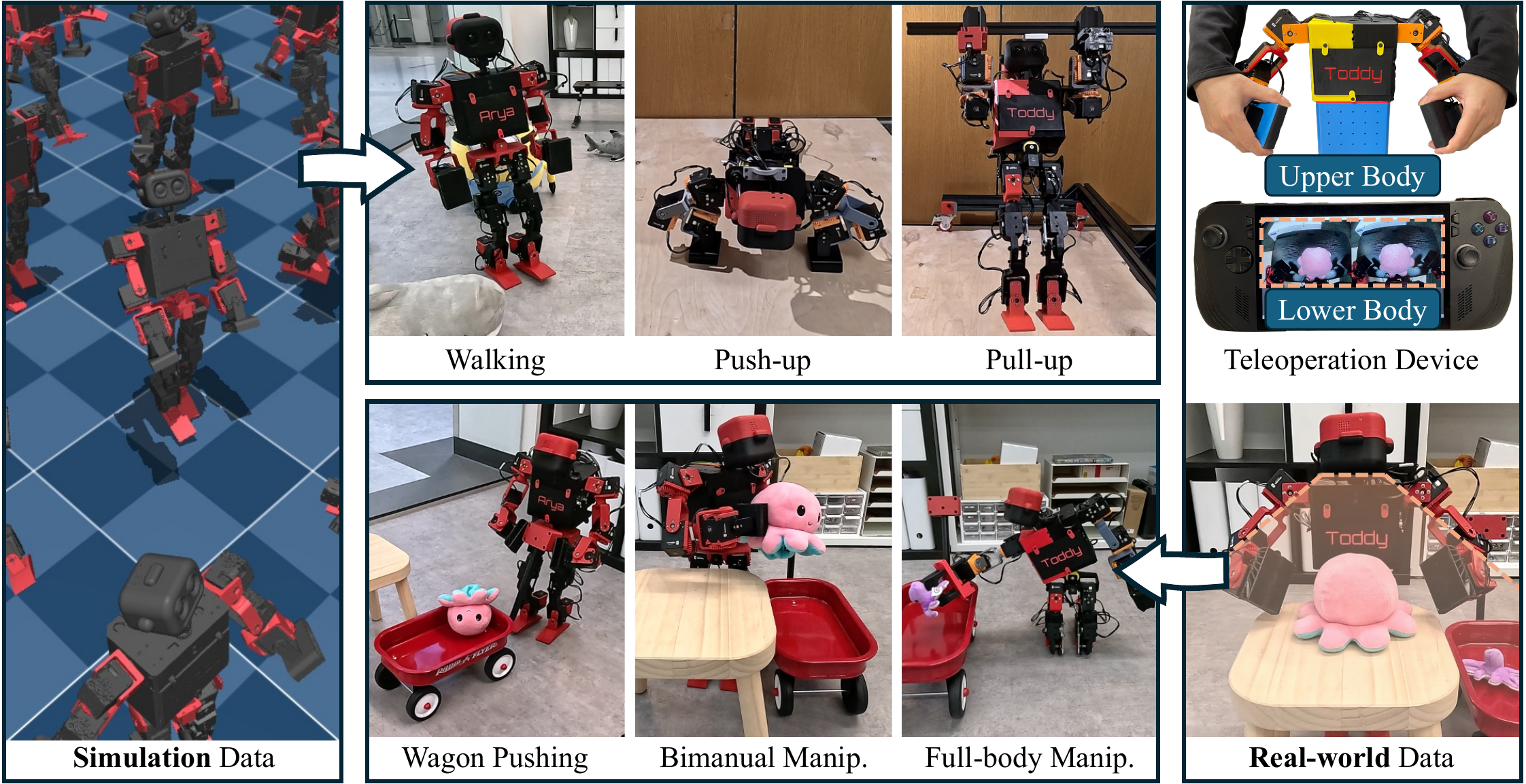}
\caption{
\textbf{\system} is an open-source humanoid platform for large-scale, high-quality data collection in both simulation and the real world. It combines massive parallel simulation and an accurate digital twin for simulation, and an intuitive teleoperation device for whole-body control in the real world. \system enables diverse loco-manipulation skills, including walking, push-ups, pull-ups, wagon pushing, bimanual, and full-body manipulation, learned from both data sources.}
\vspace{-6mm}
\label{fig:teaser}
\end{figure}


\section{Introduction}
\label{sec:intro}



Conventional robot design prioritizes actuator strength, sensor accuracy, and mechanical precision. While these characteristics are essential for deploying robots in real-world environments, they fall short in addressing the needs of robotics research—where affordability, rapid repairability, and full-stack ownership with no black boxes are crucial for faster innovation, easier debugging, and total system understanding. Additionally, traditional robot platforms are often not aligned with modern machine learning (ML) paradigms driven by embodied data. A robot platform compatible with ML must possess the innate ability to collect observation and action data, both in simulation and the real world, as these complementary data sources are essential for scalable policy learning.

In response to these challenges, we introduce \textbf{\system, an open-source humanoid robot platform} for the robotics and AI research community (Figure~\ref{fig:teaser}), specifically developed to facilitate policy learning for locomotion and manipulation skills. \system is designed to resemble a miniature human to minimize construction and maintenance costs and specialized expertise, lowering the barrier for researchers seeking to build self-sufficient, fully transparent humanoids free from critical reliance on third-party components. Although \system lacks the payload capacity of full-sized humanoids, it offers more degrees of freedom in both the upper and lower body than any existing miniature humanoids—sufficient for advancing research in manipulation and locomotion.

Inspired by recent quadrupeds~\cite{katz2019mini, kau2022stanford} and robotic manipulators~\cite{zhao2023learning, wu2023gello,shaw2023leap, romero2024eyesight, bhirangi2023all}, \system is designed with ML-compatibility in mind. Researchers can use \system to not only execute policies but also to serve as a robust data collection platform.
It enables the acquisition of \textbf{high-quality simulation data} through a plug-and-play zero-point calibration procedure and transferable motor system identification (sysID) results. These tools ensure a high-fidelity digital twin. We validate the quality of simulation data with keyframe-interpolated motions (e.g., push-ups and pull-ups) and reinforcement learning (RL) policies (e.g., walking), demonstrating the capability for zero-shot sim-to-real transfer. \system can also acquire \textbf{scalable real-world data}. We design an intuitive teleoperation interface that allows simultaneous control of \systems upper and lower body to collect whole-body manipulation data and develop effective visuomotor policies. Additionally, \systems small size and weight ($0.56~\mathrm{m}, 3.4~\mathrm{kg}$) ensure safe and accessible operation in real-world environments.

Beyond ML-compatibility, \system is designed with a focus on capability and reproducibility. A humanoid has the potential to utilize large scale human demonstrations by leveraging its anatomical similarity to the human body. As such, \system features an anthropomorphic design with 30 active degrees of freedom (excluding end effectors).
We demonstrate \systems capabilities through arm span, payload, endurance tests, and various open-loop and closed-loop loco-manipulation tasks.

Reproducibility is achieved through low-cost, open-source designs and readily accessible hardware components. \system uses commercially available motors and is completely 3D-printed, with a total cost under $6,000~\mathrm{USD}$ ($90\%$ of the cost is for motors and computers). We release digital twin, learning algorithms, hardware designs, and comprehensive tutorials to ensure that \system can be built at home with basic knowledge of hardware and software, without requiring specialized equipment for manufacturing or repair. To validate reproducibility, we enlisted a CS-major student not involved in this project to independently build another instance of \system. and successfully zero-shot transfer loco-manipulation policies between the two instances.
We also received five reports of independent replications of \system within a week by teams worldwide.

We summarize our contributions as follows: \textbf{(1)} As a system paper, we integrate existing components to create the first miniature humanoid with 30 DoFs and a superhuman range of motion, offering a practical humanoid platform with new capabilities for the community. \textbf{(2)} We develop a comprehensive sysID pipeline to ensure a high-fidelity digital twin in simulation. \textbf{(3)} We design a whole-body teleoperation solution to collect loco-manipulation data in the real world.

\section{Related Works}
\label{sec:related_works}

\begin{table*}[t]
\centering
\caption{Comparison with Other Popular Humanoid Research Platforms.}
\resizebox{\textwidth}{!}{
\begin{threeparttable}
\setlength{\tabcolsep}{3pt}

\begin{tabular}{@{}l|cccccccccc@{}}
\toprule
\textbf{Humanoid} & \textbf{Size} & \textbf{Weight} & \textbf{Compute} & \textbf{Active DoFs}$^{(b)}$ & \textbf{Manipulation} & \textbf{Locomotion} & \textbf{Sim Data} & \textbf{Real Data} & \textbf{Open Source} & \textbf{Price} \\ 
 & \textbf{(m)} & \textbf{(kg)} & \textbf{(TFLOPS)}$^{(a)}$ & \textbf{-} & \textbf{-} & \textbf{-} & \textbf{-} & \textbf{-} & \textbf{-} & \textbf{(\$)} \\ 
\midrule
BD Atlas~\citep{atlas}      & 1.50 & 89.0 & - & 28 & \tick & \tick & -   & -   & \cross        & -     \\ 
Figure~\citep{figure}        & 1.68 & 70.0 & - & 26 & \tick & \tick & -   & -   & \cross        & -     \\ 
Optimus~\citep{ai}       & 1.73 & 57.0 & - & 28 & \tick & \tick & -   & -   & \cross        & -     \\ 
Digit~\citep{agility}         & 1.75 & 65.0 & - & 16 & \tick & \tick & \tick & \cross  & code  & 250K  \\
Unitree H1~\citep{unitree}    & 1.76 & 47.0 & 1.92 & 19 & \tick & \tick & \tick & \tick & code  & 70K   \\ 
Fourier GR1~\citep{fourierrobotics}   & 1.65 & 55.0 & 2.23 & 32 & \tick & \tick & -   & -   & \cross        & 110K  \\ 
\cmidrule{0-0}
Booster T1~\citep{humanoid} & 1.18 & 30.0 & 3.33 &	23 & \tick  & \tick & \tick & \cross  & code  & 34K \\
Unitree G1~\citep{unitreea}    & 1.32 & 35.0 & 2.50 & 29 & \tick & \tick & \tick & \tick & code  & 57K   \\ 
iCub~\citep{parmiggiani2012design}          & 1.04 & 24.0 & 1.93 & 32 & \tick & \tick & \tick & \cross  & code  & 300K  \\ 
Berkeley~\citep{liao2024berkeley}      & 0.85 & 16.0 & 1.92 & 12 & \cross  & \tick & \tick & \cross  & code  & 10K   \\ 
MIT~\citep{chignoli2021mit}           & 1.04 & 24.0 & - & 18 & \cross  & \tick & -   & -   & \cross        & -     \\ 
\cmidrule{0-0}
Duke~\citep{xia2024duke}  & 1.00 & 30.0 & Optional & 10 & \cross & \tick & \tick & \cross  & design, code  & 16K  \\ 
Berkeley Lite~\citep{chi2025demonstrating}      & 0.80 & 16.0 & 0.29 & 22 & \tick  & \tick & \tick & \tick  & design, code  & 5K   \\ 
BRUCE~\citep{liu2022design}         & 0.70 & 4.8 & 0.1  & 16 & \cross  & \tick & \tick & \cross  & code  & 6.5K  \\ 
NAO H25~\citep{nao}       & 0.57 & 5.2 & 0.02  & 23 & \tick & \tick & \tick & \cross  & code  & 14K   \\ 
Robotis OP3~\citep{nameintroduction}   & 0.51 & 3.5 & 0.30  & 20 & \cross  & \tick & \tick & \cross  & code  & 11K   \\ 
Zeroth~\citep{zeroth}    & 0.48 & 3.6 & 0.01 & 16 & \tick & \tick & \tick & \cross & design, code & 1.4K  \\ 
\midrule
Ours                   & 0.56 & 3.4 & 2.50  & 30 & \tick & \tick & \tick & \tick & design, code & 6K    \\ 
Average Adult~\citep{grimmer2020human} & 1.73 & 70.9 & - & 32$^{(c)}$ & \tick & \tick & -   & -   & -        & -     \\ 
\bottomrule
\end{tabular}

\begin{tablenotes}
\item[{(a)}] TFLOPS refers to Tera FLoating-point Operations Per Second with single precision (FP32).
\item[{(b)}] The active degrees of freedom actuated by motors, excluding end effectors such as parallel grippers or dexterous hands.
\item[{(c)}] While human body is powered by over 600 muscles, the primary functional movements can be approximated using 32 revolute joints: six DoFs per leg, seven DoFs per arm, three DoFs for the waist, and three DoFs for the neck, excluding fingers and toes.
\end{tablenotes}

\end{threeparttable}
}
\label{tab:comparison}
\vspace{-4mm}
\end{table*}

In recent years, numerous humanoid robots have been developed. Full size industrial humanoids such as Boston Dynamics Atlas~\citep{atlas}, Cassie~\citep{cassie}, Digit~\citep{agility}, Figure~\citep{figure}, Tesla Optimus~\citep{ai}, Unitree H1~\citep{unitree} and Fourier GR1~\citep{fourierrobotics}. Half-scaled Booster T1~\citep{humanoid}, Berkeley Humanoid Lite ~\citep{chi2025demonstrating} and Unitree G1~\citep{unitreea}. Miniature-sized NAO H25~\citep{nao}, Robotis OP3~\citep{nameintroduction}, and K-Scale Zeroth~\citep{zeroth}. Together, they showcase diverse designs and capabilities. 
On the other hand, humanoids from research institutions, including the Berkeley Humanoid~\citep{liao2024berkeley}, BRUCE~\citep{liu2022design}, Duke Humanoid~\citep{xia2024duke}, iCub~\citep{parmiggiani2012design}, and MIT Humanoid~\citep{chignoli2021mit}, also explore the humanoid design space with different optimization emphasis.
As outlined in Table~\ref{tab:comparison}, we identify nine metrics to compare different humanoid platforms.

Size and weight are critical when designing humanoid robots. For manipulation, full-size humanoids are compatible with human-scale objects, whereas reduced-size robots are limited in their ability to manipulate them. However, reduced-size humanoids can still benefit manipulation research when paired with appropriately scaled objects~\citep{chi2025demonstrating}. Furthermore, dynamic whole-body control techniques are likely transferable to larger humanoids. For locomotion, full-sized and small-scale humanoids offer similar research values~\citep{haarnoja2024learning}.
Full-size humanoids typically require a substantial engineering team to operate and maintain, along with specialized facilities like gantry cranes for safety. In contrast, a smaller humanoid is inherently cheaper, easier to build and repair, and safer~\citep{filippini2021improving, haarnoja2024learning, yu2019sim}. They can be deployed by a single person and operated in constrained environments with a laptop.

The number of active DoFs is crucial for human-like motion. While the human musculoskeletal system employs over 600 muscles, primary functional movement can be approximated using 32 revolute joints~\citep{parmiggiani2012design, maldonado2018coordination}: six DoFs per leg, seven DoFs per arm, three DoFs for the waist, and three DoFs for the neck, excluding fingers and toes. Therefore, humanoid designs aim to achieve a DoF count as close to 32 as possible. The public perception of the limited performance of miniature humanoids is primarily due to fewer DoFs. This limitation often arises from space constraints that restrict the incorporation of many DoFs, a challenge we have successfully addressed in \system. 

To qualitatively assess the capability, we evaluate the humanoid’s ability to perform both manipulation and locomotion tasks. The combination of both is particularly compelling, as it unlocks opportunities for whole-body control research~\citep{he2024omnih2oa, he2024hover, fu2024humanplus, ji2024exbody2, lu2024mobiletelevision}. Furthermore, certain motions, such as push-ups, pull-ups, and cartwheels, require coordinated use of both arms and legs. Apart from physical capabilities, \system also has best compute in class. Onboard CUDA accelerator allows concurrent policy inferences to support visual and locomotion capabilities.
Large-scale simulation data collection is effective for locomotion~\citep{rudin2022learning, tan2018simtoreal, lee2020learning}, while real-world data collection is more promising for manipulation~\citep{oneill2024open, khazatsky2024droid, mandlekar2019scaling}. Therefore, to be \textbf{ML-compatible}, a humanoid research platform should facilitate data collection in both world.

Moreover, being open-source and low-cost is essential for others to reproduce. Without these qualities, research in this field would remain restricted to those with specialized expertise and significant resources. While making no compromise in functionality, \system stands out as completely open-source and the most accessible among recent humanoid research platforms.


\section{System Design}
\label{sec:design}

The design space of humanoid robots is vast; \systems key design principles prioritize reproducibility as a hard constraint, and capability and ML-compatibility as key design objectives.

\subsection{Reproducibility - A Hard Constraint} 
Unlike most prior works, we treat reproducibility as a hard constraint, as our humanoid platform holds no value if others cannot reproduce it. We define reproducibility as the ability of a single person to replicate the robot system at home without specialized equipment.
3D printing has become a popular method for reproducing open-source hardware systems due to its accessibility and fast turnaround time~\citep{kau2022stanford, shaw2023leap, wu2023gello}. Despite these merits, it's challenging to ensure sufficient strength and precision for printed plastic parts. We discuss our justifications for 3D printing in Appendix~\ref{sec:3d_printing}. For performance-critical components like motors and bearings, we limit ourselves to off-the-shelf parts.
The total BOM cost of \system is $6,000~\mathrm{USD}$, with 90\% spent on the computer (NVIDIA Jetson) and motors (Robotis Dynamixel). We detail our design considerations in Appendix~\ref{sec:ease_to_build}.

\begin{figure*}[t]
  \centering
  \includegraphics[width=0.98\textwidth]{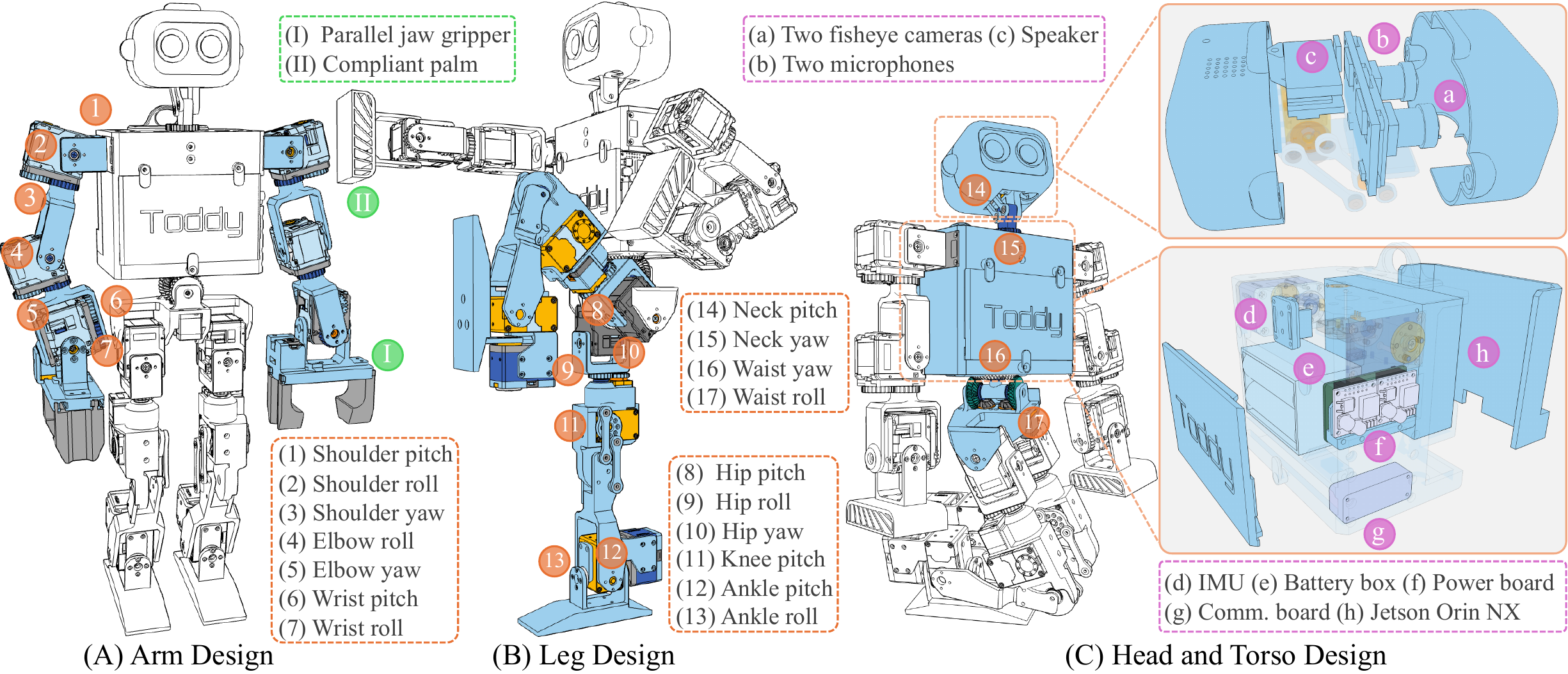}
  \caption{\textbf{Mechatronic Design.} 
  Orange markers highlights \systems 30 active DoFs: 7 per arm, 6 per leg, a 2 on neck, and a 2 on waist. Green markers indicate two end-effector designs—a compliant palm and a parallel-jaw gripper. Purple markers denote the electronics layout with exploded views, featuring 2 fisheye cameras, 1 speaker, 2 microphones, 1 IMU, 
  and 1 Jetson Orin computer.}
  \label{fig:design}
  \vspace{-3mm}
\end{figure*}


\subsection{Capability - Design Objective}

\textbf{Mechatronic Design.} As shown in Figure~\ref{fig:design}, \system replicates the structure of an adult human body to maximize the number of active DoFs, enabling human-like motion and versatile loco-manipulation tasks. Each arm features seven DoFs with spur gears for axis-aligned transmission, ensuring high functionality and reachability. Each leg has six DoFs: three at the hip for a wide range of motion and optimized walking strides, a parallel linkage at the knee to reduce inertia, and two at the ankle for stable locomotion. The neck, with two DoFs, incorporates parallel linkages at the pitch joint for a compact design, allowing expressive motion and full head mobility. The waist, also with two DoFs, uses coupled bevel gears to balance the space budget and effectively transmit power from two motors for yaw and roll actuation, which enables whole-body control. Additionally, \system offers two end-effector designs: a parallel jaw gripper for grasping and a compliant palm for tasks requiring a palm-like posture. The end-effector designs can be switched quickly within two minutes. 
For each active DoF, we maximize the range of motion by optimizing geometries to prevent self-collisions (Appendix~\ref{sec:range_of_motion}). To address concerns such as space constraints, axis alignment, and inertia reduction, we integrated a combination of three transmission mechanisms in the design: spur gears, coupled bevel gears, and parallel linkages (Appendix~\ref{sec:transmission}). 


\textbf{Sensors, Compute, and Power.} To enhance \systems capabilities, we integrate a comprehensive set of sensors and computational components. Two fisheye cameras are included to expand the field of view. An Inertial Measurement Unit (IMU) is placed in the chest to provide state feedback. A speaker and microphones facilitate communication with humans and other robots. We provide a conversation example between two \system instances in the supplementary video.
The onboard computation is powered by a Jetson Orin NX 16GB, enabling real-time inference of ML models. 
Based on our proposed power metric (Appendix~\ref{sec:power_metric}), we choose Dynamixel motors for their reliability and accessibility (Appendix~\ref{sec:motor_selection}). Power management is handled by a custom-designed power distribution board which can be easily ordered online (Appendix~\ref{sec:power_budget}). 


\subsection{ML-Compatibility - Design Objective}

\textbf{Digital Twin.} A high-fidelity digital twin is fundamental for high-quality simulation data collection and zero-shot sim-to-real transfer.
We divide the digital twin development into two key components: zero-point calibration for correct kinematics and motor system identification for accurate dynamics.
Since Dynamixel motors lack an absolute zero point, a reliable method is needed to calibrate after reassembly. We designed a set of calibration devices that can align the robot to the desired zero point within a minute, defined as standing with both arms parallel to the body (Appendix~\ref{sec:calibration}).
For accurate sysID, we designed a motor test bed and automatic test procedures described in Appendix~\ref{sec:motor_test_bed} to measure the range of dynamics parameters.
Next, we collect actuator position tracking data by commanding a chirp signal~\citep{haarnoja2024learning},  and use the position tracking result to fit an actuation model~\citep{grandia2024design} with optimization, as detailed in Appendix~\ref{sec:actuation_model}. 
Testing across instances shows Dynamixel motors of the same model have nearly identical dynamics parameters. This is also validated by successful direct deployment of policies to a second \system instance without additional sysID.

\textbf{Teleoperation Device.}  
Inspired by previous works~\citep{wu2023gello, aldacoaloha}, we develop a second upper body of \system as the leader arms to collect high-quality real-world data. Two force-sensitive resistors (FSRs) are embedded in the gripping area of the end effectors to detect compression force from the operator, allowing gripper movement based on force input.
We use a handheld gaming computer (either Steam Deck or ROG Ally X) to control the other body parts. The joysticks send velocity commands to walk, turn, and squat. Buttons trigger either programmed or trained policies
and provide direct control over neck and waist movements. The detailed mapping is described in Appendix~\ref{sec:joystick}.
\section{System Control}
\label{sec:control}

\subsection{Keyframe Animation}

Keyframe animation is widely used in character animation, but it provides only kinematics data, lacking a guarantee of dynamic feasibility~\citep{izani2003keyframe}. To address this, we develop a tool integrating MuJoCo~\citep{todorov2012mujoco} with a GUI, enabling real-time tuning and validation of keyframe motion trajectories.
Combined with high-fidelity digital twin, we can efficiently generate open-loop trajectories such as cuddling, push-ups, and pull-ups that can be executed zero-shot in the real world.

\subsection{Reinforcement Learning}


For walking, we train a reinforcement learning (RL) policy, $\pi(\bm{\mathrm{a}}_t | \bm{\mathrm{s}}_t)$, which outputs $\bm{\mathrm{a}}_t$ as joint position setpoints for proportional-derivative (PD) controllers, based on the observable state $\bm{\mathrm{s}}_t$:
\begin{equation}
\bm{\mathrm{s}}_t = \left(\bm{\phi}_t, \bm{\mathrm{c}}_t, \Delta\bm{\mathrm{q}}_t, \bm{\dot{\mathrm{q}}}_t, \bm{\mathrm{a}}_{t-1}, \bm{\theta}_t, \bm{\omega}_t \right),
\end{equation}
where $\bm{\phi}_t$ is a phase signal, $\bm{c}_t$ represents velocity commands, $\Delta\bm{q}_t$ denotes the position offset relative to the neutral pose $\bm{q}_0$, $\bm{a}_{t-1}$ is the action from the previous time step, $\bm{\theta}_t$ represents the torso orientation, and $\bm{\omega}_t$ is the torso’s angular velocity. 
%
During PPO policy training~\citep{schulman2017proximal}, the environment generates the next state, $\bm{s}_{t+1}$, updates the phase signal, and returns a scalar reward $\mathrm{r}_t = \mathrm{r}(\bm{s}_t, \bm{a}_t, \bm{s}_{t+1}, \bm{\phi}_t, \bm{c}_t)$. Following standard practice~\citep{grandia2024design}, the reward is decomposed as:
\begin{equation}
    \mathrm{r}_t = \mathrm{r}_t^{\text{imitation}} + \mathrm{r}_t^{\text{regularization}} + \mathrm{r}_t^{\text{survival}}.
\end{equation}

Among these components, $\mathrm{r}_t^{\text{imitation}}$ encourages accurate imitation of the reference walking motion, which is generated using a closed-form ZMP (Zero Moment Point) solution~\citep{tedrake2015closedform}.  $\mathrm{r}_t^{\text{regularization}}$ incorporates heuristics of ideal walking motion, penalizes joint torques, and promotes smooth actions to minimize unnecessary movements. A survival reward $\mathrm{r}_t^{\text{survival}}$ prevents early episode termination during training. Additional details are provided in Appendix~\ref{sec:rl_details}.

\subsection{Imitation Learning}

Real-world data collection involves a human operator guiding the leader’s arms to teleoperate the follower’s arms while using a joystick and buttons on a handheld game controller to control body movements.
During data collection, when the upper body tracks the position commands from the leader arms, \systems lower body employs a two-layer PD controller to actively maintain balance. The first layer is a Center of Mass (CoM) PD controller, which keeps the CoM close to the center of the support polygon. The second layer is a torso pitch PD controller, which uses IMU readings to ensure the torso remains upright. The first layer addresses CoM shifts caused by arm movements, while the second layer compensates when lifting heavy objects.
With this setup, we can collect 60 trajectories in just 20 minutes for both bimanual and full-body manipulation tasks. The motor positions of the leader's arms are recorded as the actions, while the motor positions of the follower robot, along with the RGB images captured from its camera, are recorded as observations. We train a diffusion policy~\citep{chi2023diffusion} on this dataset, with further details provided in Appendix~\ref{sec:dp_details}.

\section{Experiments}
\label{sec:experiments}

\begin{figure}[t]
  \centering
  \includegraphics[width=\linewidth]{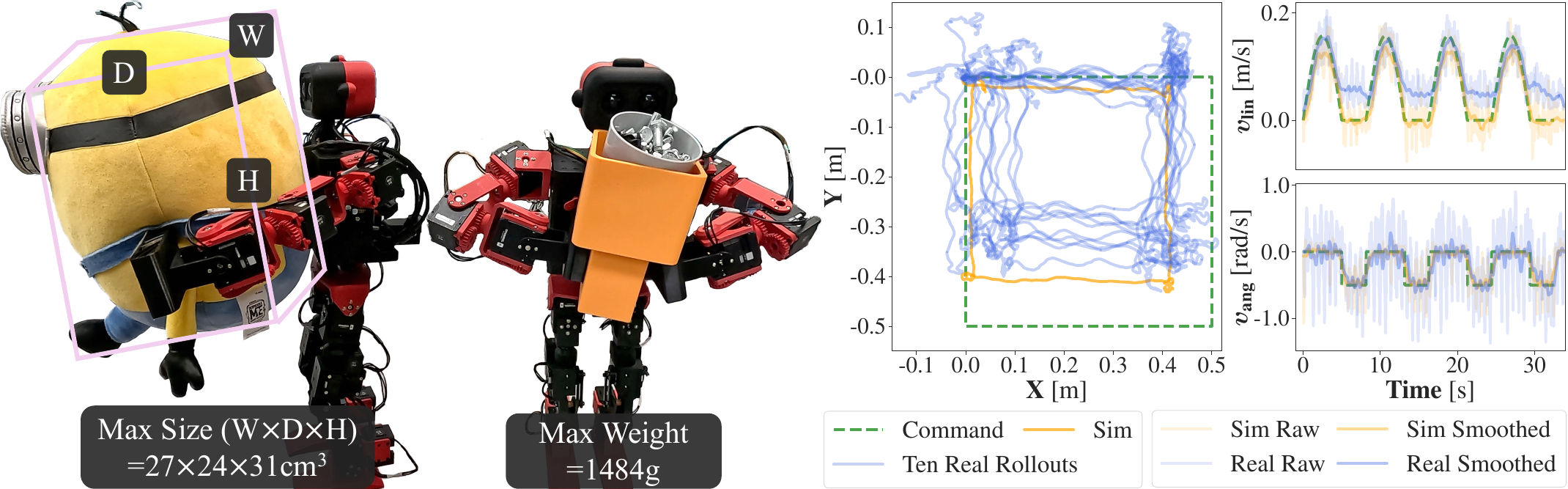}
  \caption{\textbf{Arm Span, Payload, and Trajectory Tracking.} On the left, we show that with a torso dimension of $13 \times 9 \times 12~\mathrm{cm}^3$, \system can grasp objects up to $27 \times 24 \times 31~\mathrm{cm}^3$, about 14 times the torso size. Additionally, \system can lift weights up to $1484~\mathrm{g}$, which is 40\% of its body weight ($3484~\mathrm{g}$). On the right, we present ten consecutive real-world rollouts of an RL walking policy tracking a square trajectory with a predefined velocity profile.
    Both raw and smoothed linear and angular velocity tracking are displayed, with real-world results averaged across trials.}
  \label{fig:hardware_eval}
    \vspace{-3mm}
\end{figure}


\textbf{Capability: Arm Span, Payload, and Endurance.}
To evaluate \systems arm span, we teleoperate it to hug a large object using the compliant palm gripper while maintaining balance. We show that, \system can grasp objects up to $27 \times 24 \times 31~\mathrm{cm}^3$, approximately 14 times its torso volume. 

The payload test assesses both the upper body’s lifting capacity and the lower body’s ability to maintain balance. \system successfully lifts up to $1484~\mathrm{g}$, 40\% of its total weight ($3484~\mathrm{g}$). 
To eliminate friction effects, we use a 3D-printed cup for the gripper to lock in securely. We incrementally add screws to the cup until \system falls over. Results are shown on the left of Figure~\ref{fig:hardware_eval}.

In the endurance test, \system starts with a fully charged battery, running the walking RL policy while stepping in place. \system achieved the longest streak of 19 minutes without falling. Over time, increased motor temperatures gradually pushed it outside the policy’s training distribution, leading to more frequent falls. \system withstands up to 7 falls before breaking, but even then, repairs are quick—requiring only 21 minutes of 3D printing and 14 minutes of assembly, including removing the damaged part, installing the replacement, and performing zero-point calibration.

\begin{figure*}[t]
  \centering
  \includegraphics[width=0.98\textwidth]{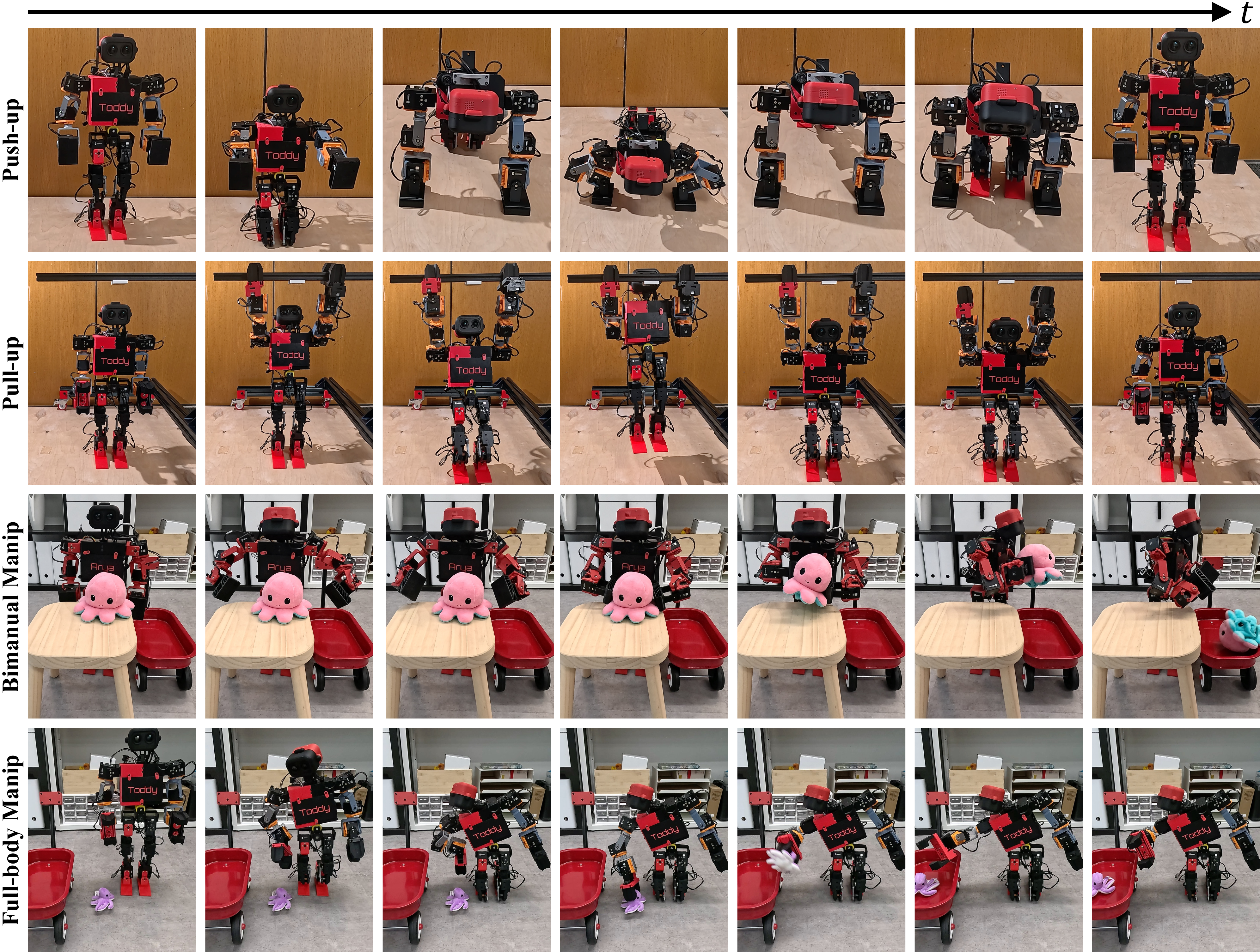}
  \caption{\textbf{Experiment Results.} We present four different tasks: push-up, pull-up, bimanual, and full-body manipulation, showing \systems capability in challenging loco-manipulation tasks.}
  \label{fig:results}
  \vspace{-5mm}
\end{figure*}

\textbf{Capability: Push-ups and Pull-ups.}
To demonstrate expressive and dynamic motions, we program push-ups and pull-ups in our keyframe software and perform zero-shot sim-to-real transfer (Figure~\ref{fig:results}). For pull-ups, we use an AprilTag to help \system locate the horizontal bar. Both tasks require strong limbs, balanced upper-lower body strength, and precise coordination, particularly when \system transitions from a planking pose to standing after push-ups and when it releases the horizontal bar and lands after pull-ups. These open-loop transfers require only a single motion trajectory designed in simulation, highlighting the fidelity of our digital twin. 

\begin{figure*}[t]
  \centering
  \includegraphics[width=\linewidth]{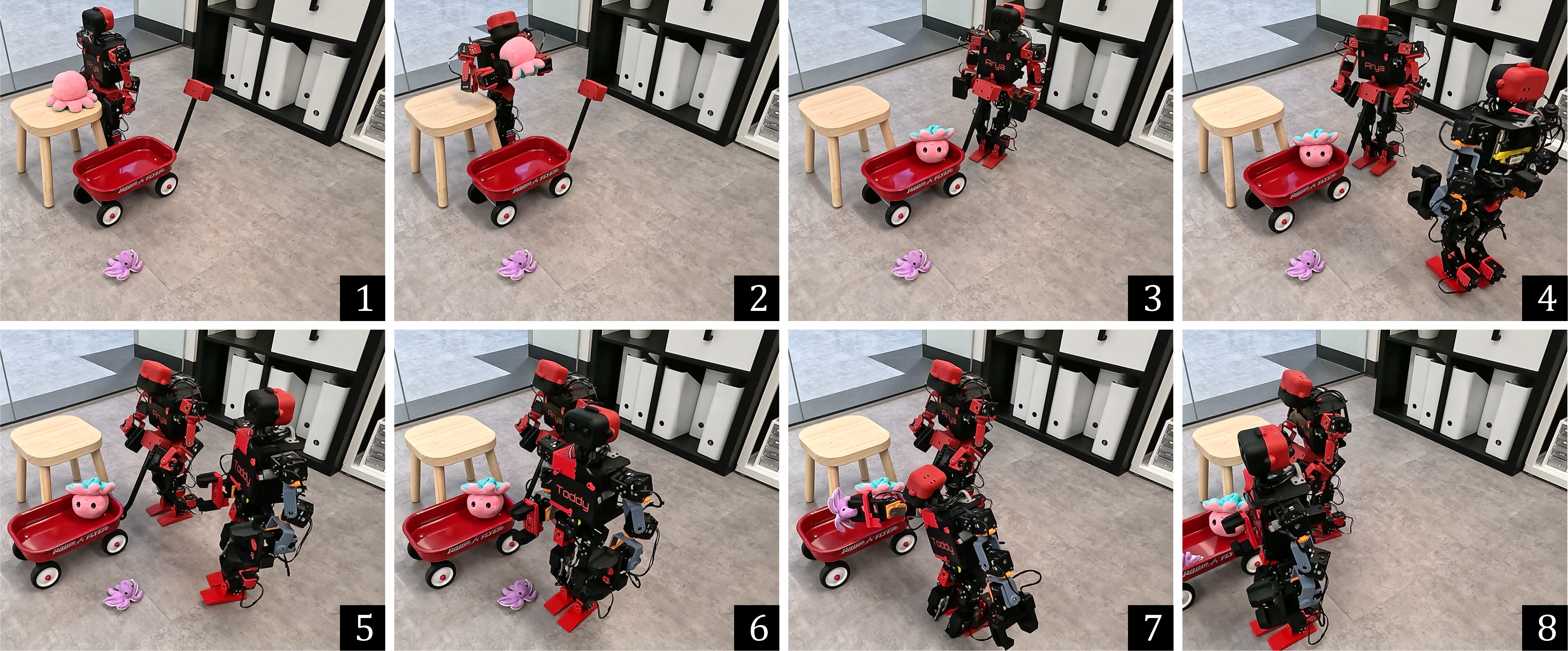}
  \caption{\textbf{Long-horizon Collaboration}. In this task, two instances of \system, Arya and Toddy, collaborate to clean up a toy session. (1) The task begins with a pink octopus on the table and a purple octopus on the ground. (2) Arya picks up the pink octopus from the table and places it in the wagon. (3) Arya walks to the wagon handle. (4) Arya grasps the handle while Toddy walks over. (5) Arya pushes the wagon toward the purple octopus. (6) Toddy reaches the pickup position. (7) Toddy kneels and picks up the purple octopus. (8) Finally, Arya and Toddy leave side by side.}
  \label{fig:collaboration}
  \vspace{-3mm}
\end{figure*}

\textbf{ML-Compatibility: Omnidirectional Walking.}
To demonstrate \systems locomotion capabilities, we train RL walking policies to follow a square trajectory with a predefined velocity profile. The right part of Figure~\ref{fig:hardware_eval} presents the results, with real-world tracking data collected via motion capture. Due to RL policy limitations, both simulation and real-world tracking deviate from the command, primarily because the learned walking policy struggles with in-place rotation, causing translation offsets. However, the sim-to-real gap is notably smaller than the tracking gap, supporting a successful zero-shot transfer. Additionally, we report a position tracking error variance of $0.018~\mathrm{m}$, which demonstrates good repeatability (more details in Appendix~\ref{sec:vel_tracking}).

\textbf{ML-Compatibility: Vision-based Manipulation.}
We show \systems ability to perform bimanual manipulation and full-body manipulation by transferring octopus toys from a table and the ground to a wagon. Both tasks are trained with RGB-based diffusion policy~\citep{chi2023diffusion} with 60 demonstrations. Across 20 test trials, we achieve a 90\% success rate for bimanual manipulation and 75\% for full-body manipulation. We leverage a combination of open-loop motions and closed-loop policies to enhance data collection efficiency. In the bimanual task, the torso rotating and releasing motions are open-loop, while in full-body manipulation, kneeling down is open-loop. 
\systems onboard computing runs a $300$M parameter diffusion policy with about $100~\mathrm{ms}$ latency, enabling real-time operation. 
Results are shown in Figure~\ref{fig:results}.

\textbf{ML-Compatibility: Skill Chaining}
We test \systems ability to combine loco-manipulation skills with wagon pushing. The results are shown in Appendix~\ref{sec:skill_chaining}. 
To push the wagon, \system first executes a diffusion policy to grasp the handle, while maintaining that pose, switches to the RL policy to walk forward. To enable walking while maintaining the grip, we sample the robot's end pose from the grasping policy’s training data (60 demonstrations) during RL training. 

\textbf{Reproducibility: Hardware and Policies}
To demonstrate hardware reproducibility, we recruit a CS-major student with no prior hardware experience to assemble a second \system using our open-source assembly manual and videos. The student independently completes the assembly in three days, including the time for 3D printing. The open-source community has also reported successful reproductions of \system, and most of them were completed in one week (Appendix~\ref{sec:reproduction}).
For policy reproducibility, we run the manipulation policy trained on data collected with one instance on the other instance, achieving the same success rate of 90\% across 20 trials. We also successfully transferred the RL walking policy between both robots. To further showcase the equivalent performance of both \system instances, we have them collaborate on a long-horizon toy tidy-up session, as shown in Figure~\ref{fig:collaboration}.



\section{Conclusions}
\label{sec:conclusion}

In conclusion, we demonstrate that \system is ML-compatible, capable, and reproducible through a series of tests and loco-manipulation tasks. \system extends beyond locomotion and supports full-body manipulation, character animation, human-robot interaction, and a range of ML applications, making it a versatile research platform. To bridge the sim-to-real gap, we develop a comprehensive sysID pipeline to create a high-fidelity digital twin, alongside a whole-body teleoperation system to collect loco-manipulation data. Being fully open-source, \system empowers researchers to explore new research directions and fosters open collaboration across the community.

\section{Limitations and Future Work}
\label{sec:limitation}

\systems performance in more agile tasks is constrained by the off-the-shelf motors' max speed, max torque, and communication speed. Should there be stronger motor choices, the performance can still improve. Rather than achieving superhuman capabilities, \system aligns more closely with average human performance in loco-manipulation tasks. In addition, the current actuation model does not consider motor temperature and is less accurate when close to the performance limit. The miniature scale limits its interaction with human-sized objects, though this does not hinder research if appropriately sized objects are used. Compared to metal shells, 3D-printed parts are more likely to break after the event of impact. Despite 3D printing being fast, repairs still require time.

To overcome these limitations, we plan to develop customized communication boards to increase control frequency and improve our actuation model to maximize usable envelope, especially when the motor heats up. We will refine \systems design to improve structural strength, as we believe that design has a greater impact than material choice. We also aim to improve sensing capabilities, including stereo vision for depth perception, additional IMUs for improved state estimation, and tactile sensors for richer manipulation feedback. 
	

\acknowledgments{
The authors would like to express their gratitude to Kaizhe Hu for assembling the second instance of ToddlerBot and assisting with keyframe animation and demo recording. We also extend our thanks to Huy Ha, Yen-Jen Wang, Pei Xu, and Yifan Hou for their insightful discussions on locomotion, and to Sirui Chen, Chen Wang, and Yunfan Jiang for valuable input on manipulation policy deployment.
We are grateful to Albert Wu for his guidance on mathematical formulation and notation. Additionally, we thank Jo\~{a}o Pedro Ara\'{u}jo for his assistance with the motion capture system.
Finally, we appreciate the helpful discussions from all members of TML and REALab. 
This work was supported by National Science Foundation NSF-FRR-2153854, NSF-2143601, NSF-2037101, NSF-2132519, Sloan Fellowship, Stanford Institute for Human-Centered Artificial Intelligence, and Stanford Wu Tsai Human Performance Alliance.
}

\clearpage

\bibliography{references}  

\newpage

\section{Appendix}
\label{sec:supplementary}

\subsection{3D Printing}
\label{sec:3d_printing}
3D printing is a deliberate design choice when combined with miniature-sized parts, not just a cost-saving measure. A quick derivation from Euler-Bernoulli beam bending theory suggests that ToddlerBot's smaller structures experience much less relative deflection with the same material strength---$\delta = \frac{PL^3}{3EI} \rightarrow \frac{\delta}{L}\propto \frac{PL^2}{3EL^4}=\frac{P}{3EL^2}$, where $\delta$ is deflection, $P$ is point load at tip, $E$ is material elastic modulus, $L$ is characteristic length, and $I\propto L^4$ is second moment of area. We estimate that our miniature (smaller L) 3D-printed parts (smaller E) offer comparable strength to full-size aluminum ones. Using metal parts would reduce reproducibility with marginal gains. Our gears are carefully tuned for minimal backlash, precise transmission, and direct motor integration, while off-the-shelf parts can’t achieve this. We also incorporate various bearings to minimize friction in power transmission components.
After over a year of testing on two ToddlerBot instances, we are confident that ToddlerBot’s reliability matches that of commercial platforms. 
\subsection{Ease of Assembly and Maintenance}
\label{sec:ease_to_build}
The ease of assembly and maintenance is crucial yet difficult to optimize, as it requires mentally simulating the assembly and disassembly process. In early iterations, we explicitly optimize for fewer screw types and ensured unobstructed tool access, which allowed a clear assembly direction for the screwdriver. We also prioritized modular design, allowing individual parts to be replaced independently. These considerations significantly improve maintainability and simplify repairs. As we continue our research on this platform, we have addressed several reliability issues, including motor overheating and link durability. We are committed to releasing ongoing fixes and upgrades to improve the humanoid further.

\subsection{Range of Motion}
\label{sec:range_of_motion}

\begin{figure}[h]
  \centering
  \includegraphics[width=0.7\linewidth]{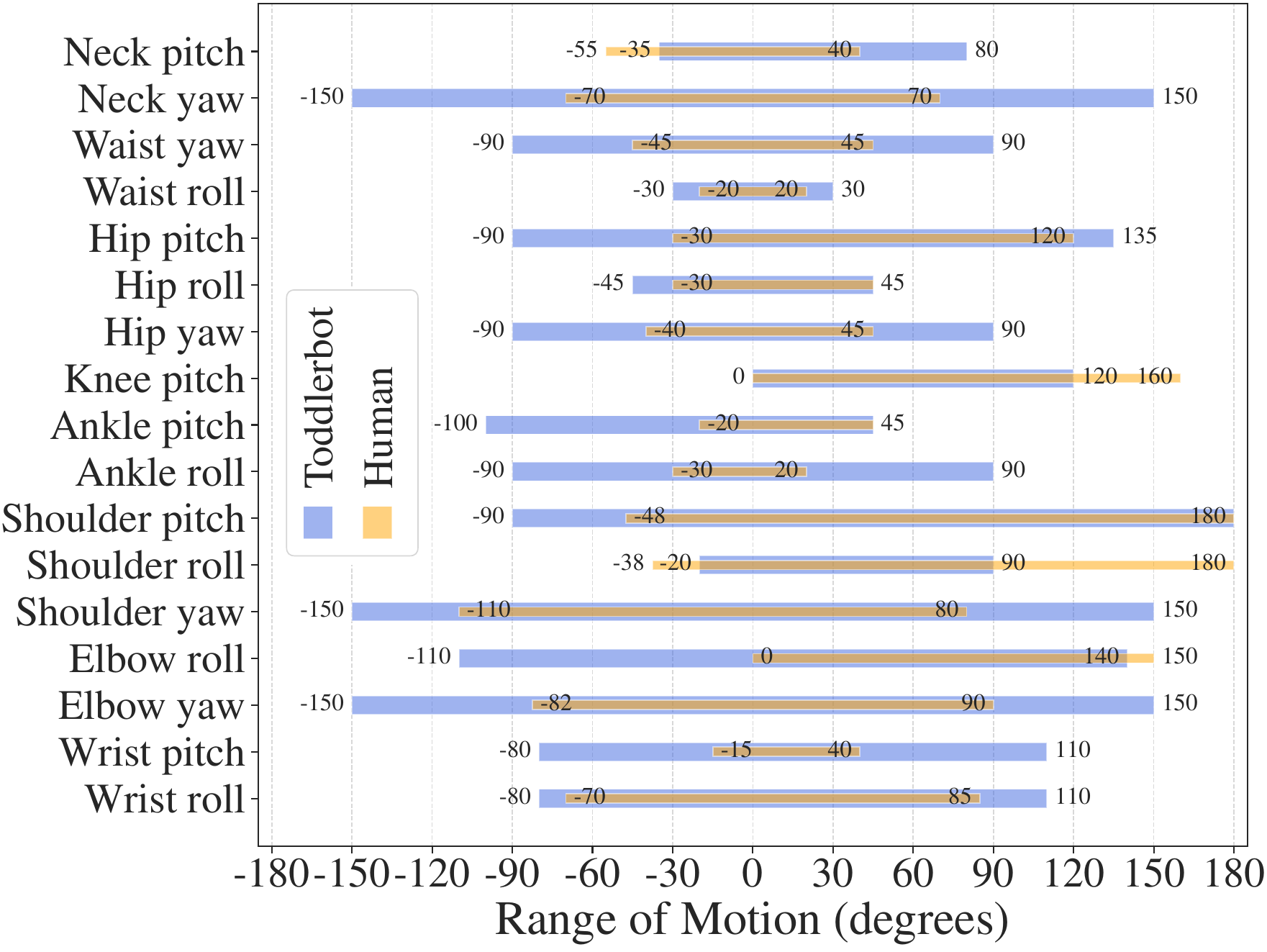}
  \caption{\textbf{Range of Motion.} We show that \system has near-human or even superhuman mobility in most joints. Negative values represent extensions, adductions, and inversions, while flexions, abductions, and eversions are positive.}
  \label{fig:rom}
  \vspace{-2mm}
\end{figure}

We design each joint’s range of motion based on human biomechanics, optimizing geometries to prevent self-collisions and achieve near-human or even superhuman mobility in most joints (Figure~\ref{fig:rom}).
The additional extension of the ankle pitch partially compensates for the $40\degree$ gap in the knee pitch and the absence of toes. For shoulder roll, although $90\degree$ is limited compared to human shoulder abduction, the same hand-up pose can be achieved by actuating the shoulder pitch joint.
\subsection{Transmission Mechanisms}
\label{sec:transmission}

\begin{figure}
  \centering
  \includegraphics[width=0.5\linewidth]{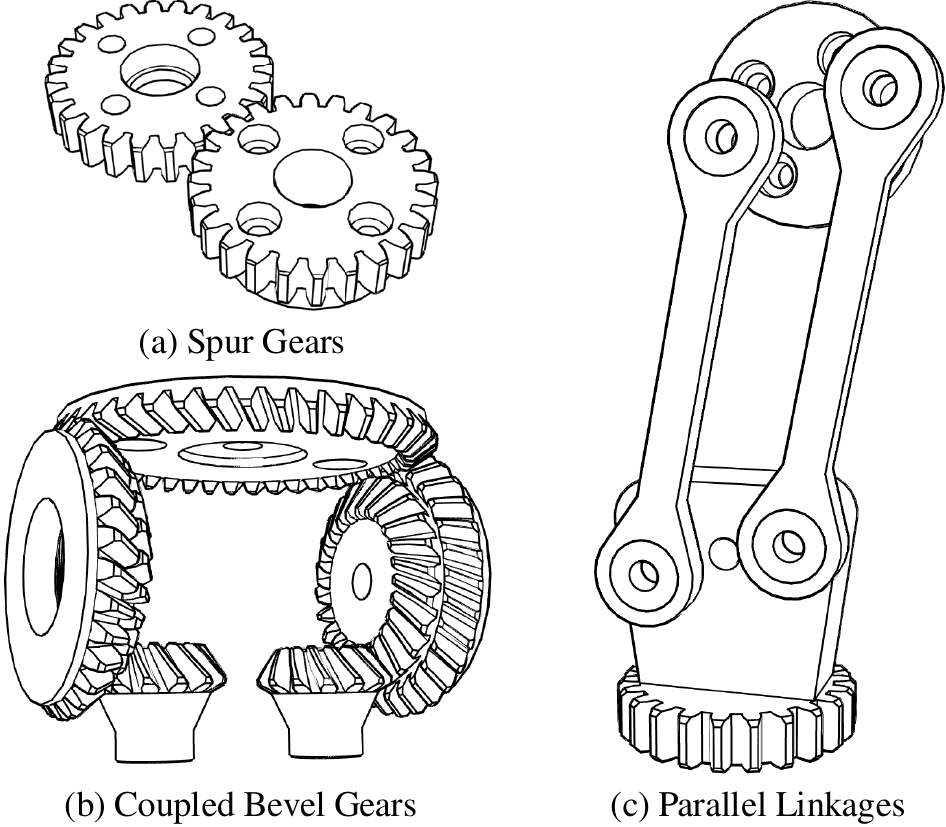}
  \caption{\textbf{Transmission Mechanisms.} We show three design primitives in \systems mechanical design: spur gears, coupled bevel gears, and parallel linkages.}
  \label{fig:transmission}
  \vspace{-4mm}
\end{figure}

Placing motors directly at the joint is often impractical due to the limited space budget. With carefully designed transmission mechanisms, motors can be relocated outside the interference zone, amplify torque output, and offload mechanical stress to the structure. This section highlights key design features, including spur gears, coupled bevel gears, and parallel linkages, as shown in Figure~\ref{fig:transmission}. 

Each transmission type offers unique benefits. To start with, spur gears provide three advantages:

\begin{itemize}[leftmargin=4mm]
    \item \textbf{Relocated joint axis:} A 1:1 spur gear set allows repositioning of the joint axis to a more convenient in-plane location. This is widely used in \systems arm.
    \item \textbf{Torque modification:} A ratioed spur gear set adjusts the final torque output, which is particularly useful for the parallel jaw gripper.
    \item \textbf{Load distribution:} When a motor’s output shaft has significant free play, as in Dynamixel XC330, where it is supported only by a Teflon bushing, using it directly as the joint axis is undesirable. A 1:1 spur gear set enables a reinforced secondary axis with planar bearings and metal shafts to carry the load, protecting the motor from transverse forces. This approach is used in the hip yaw joints, where torque demands are low, but load-bearing capacity is critical.
\end{itemize}

With precise tolerance tuning, 3D-printed bevel gears provide a highly interlocking design with minimal backlash, while still being structurally strong. They also offer three key advantages:

\begin{itemize}[leftmargin=4mm]
\item \textbf{Rotated joint axis:} A coupled bevel gear set enables a parallel waist mechanism, where two motors in the same orientation drive two perpendicular DoFs.

\item \textbf{Combined torque output: } On each axis, both motors contribute to the driving torque, enhancing power and efficiency. This is critical, as a single Dynamixel XC330 lacks the power to drive the entire upper body, but two motors combined are sufficient.

\item \textbf{Compact actuation:} In the waist, where space is highly constrained, a coupled bevel gear set allows the compact integration of two DoFs. 
\end{itemize}

Lastly, parallel linkages allow the motor to be positioned away from the joint axis, as seen in the knee and neck pitch. Despite a slightly smaller range of motion (usually $<160\degree$), this design efficiently transfers high torque when paired with ball bearings. They provide three key benefits:

\begin{itemize}[leftmargin=4mm]
\item \textbf{Compact design:} This enables a cleaner neck design by placing the motor inside the head.
\item \textbf{Reduced Inertia:} The knee motor is placed higher to reduce rotational inertia.
\item \textbf{Structural Efficiency:} In the thigh, the knee motor is bolted to a 3D-printed structure for better load distribution, increased rigidity, and reduced weight.
\end{itemize}

A potential drawback of these transmission mechanisms is their inaccurate simulation modeling. However, in MuJoCo~\citep{todorov2012mujoco}, we mitigate this by using joint equality constraints for spur gears, fixed tendons for coupled bevel gears, and weld constraints for parallel linkages. This approach has empirically shown a small sim2real gap, as demonstrated in Section~\ref{sec:experiments}.

\subsection{Power Factor}
\label{sec:power_metric}

\begin{figure}[t]
  \centering
  \includegraphics[width=0.6\columnwidth]{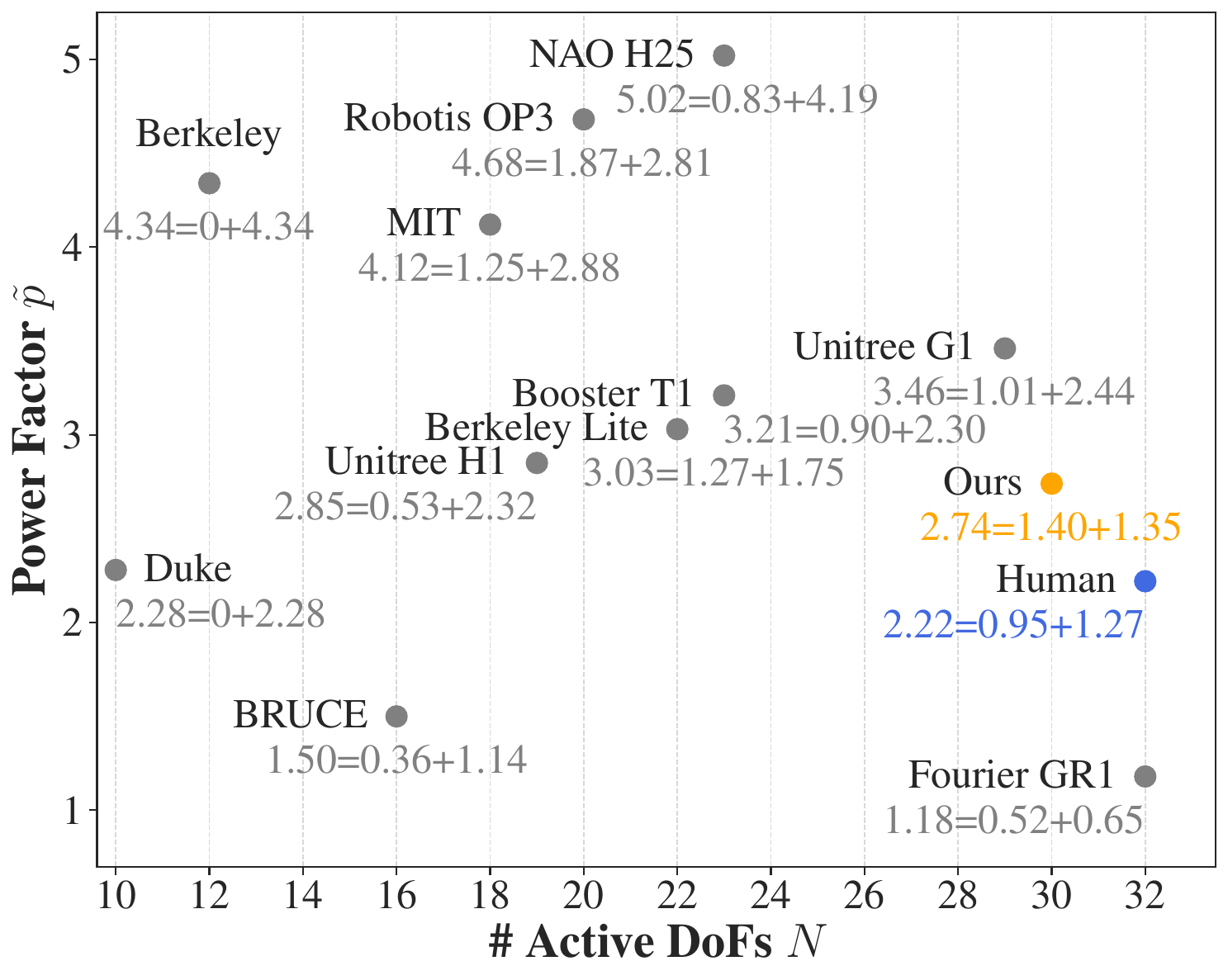}
  \caption{\textbf{Humanoid Metrics.} Two key criteria of humanoid capability are the number of active DoFs and power factor $\tilde{p}$ (Equation~\ref{eq:performance}). The total power factor is the sum of the upper and lower body power factor: $\tilde{p}_{\text{total}}=\tilde{p}_{\text{upper}}+\tilde{p}_{\text{lower}}$. \system is the closest to human compared with other humanoids, implying potentially comparable loco-manipulation capabilities.}
  \label{fig:performance}
  \vspace{-3mm}
\end{figure}

To quantitatively assess a humanoid robot’s capability, we propose
$\tilde{\mathit{p}}$ as the power factor, representing the total torque (and thus mechanical power) a robot can generate relative to its weight and height.
%
Intuitively, a higher $\tilde{\mathit{p}}$ means that a humanoid can perform energetic, dynamic motions more easily. We argue that $\tilde{\mathit{p}}$ should at least exceed the human threshold $\tilde{\mathit{p}}_{\text{human}}$ to achieve human-like motion, given the inherent gap between robot and human policies, assuming humans operate as an oracle policy that is energy efficient.
However, raising $\tilde{\mathit{p}}$ far beyond $\tilde{\mathit{p}}_{\text{human}}$ can have adverse effects: unnatural motion, excessive reliance on motor power, fewer DoFs to accommodate larger motors, reduced battery life, and increased safety concerns. Thus, pushing $\tilde{\mathit{p}}$ past diminishing returns involves a practical trade-off. As shown in Figure~\ref{fig:performance}, \system has a $\tilde{\mathit{p}}$ score closest to humans.


When comparing the performance of humanoids with different scales and weights, directly having a full-sized $1.8~\mathrm{m}$ humanoid and a $0.5~\mathrm{m}$ scaled humanoid both jump $0.5~\mathrm{m}$ or run at $3~\mathrm{m/s}$ is not a fair comparison. A more reasonable approach is to normalize performance metrics, for example, by evaluating a jump at $10\%$ of body height or a running speed of twice the body length per second. Formally, we say two humanoids have the \textbf{same performance} if they execute the same sequence of joint motions over a time span $T$, and their total power consumption is the same fraction of their motors’ maximum power:
\begin{equation}
     \frac{\int_0^T{{p}(t) dt}}{\sum_{i=0}^{N}|{\tau}_{i}^{\text{max}}\dot{q}_i|} \approx
     \frac{\Delta h \cdot mg}{\sum_{i=0}^{N}|{\tau}_{i}^{\text{max}}\dot{q}_i|} \approx   \frac{h \cdot mg}{\sum_{i=0}^{N}|{\tau}_{i}^{\text{max}}\dot{q}_i|},
\label{eqn:power_factor_explained}
\end{equation}
where $p(t)$ is the humanoid's power output at time $t$. $\tau^\text{max}_i$ and $\dot{q}_i$ are the maximum torque and joint velocity of the $i$-th motor in the humanoid respectively. Moreover, when computing the maximum power, we take the absolute value to ensure that all motors do positive work. For the numerator, the integral of the power output over $T$, which is equivalent to the work done by the robot, can be approximated by the gravitational energy gained: $\Delta h \cdot mg$. Since $\Delta h$ is approximately proportional to the height of the humanoid, we further replace $\Delta h $ with just the humanoid's height $h$.

Based on Equation \ref{eqn:power_factor_explained}, we now define power factor to measure the performance of a humanoid:
\begin{equation}
    \tilde{\mathit{p}} = \frac{\sum_{i=0}^{\mathit{N}} |\tau_{i}^{\max}|}{h \cdot m g}.
    \label{eq:performance}
\end{equation}
Note that we flip the fraction from Equation \ref{eqn:power_factor_explained} to make the power factor value increase as the utilized torque ratio decreases. We also drop $\dot{q}$ as the same sequence of joint motion would be executed when using power factor to compare the performance of humanoids.

\citet{chi2025demonstrating} proposes a similar power metric but with an additional normalization by $N$, the number of active DoFs. However, we find it to be an inappropriate modification. Consider an exaggerated case where two humanoids have the same height and weight: one have a single DoF and the other have 100 equally capable DoF. Normalizing by $N$ would assign them the same power factor, which does not reflect the true actuation capacity of the system at all.

\subsection{Motor Selection}
\label{sec:motor_selection}

We choose Dynamixel motors because of their robustness, reliability, and accessibility. Different types of Dynamixel motors were selected for various joints based on space constraints, torque requirements, and cost considerations. In terms of communication speed, Dynamixel motors communicate via a $5\mathrm{V}$ TTL protocol running at 2M baudrate, providing full-state feedback for all 30 motors at $50~\mathrm{Hz}$ using an off-the-shelf communication board. 
Precision-wise, our motors have a backlash of about 0.25\degree, on
par with most QDD joints used in full-scale humanoids. Also, the Dynamixel encoder resolution is 4096 pulse/rev, and
thus the resolution of the motor position reading is 0.09\degree.

Given \systems size constraint and 30-DoF design, Brushless Direct Drive (BLDC) motors are not a viable option. As BLDC motors shrink, their winding thickness decreases, reducing current capacity and torque constant. Despite their high power density, they still require a high-ratio gearbox, making them less suitable given our limited space budget.
We initially explored electric linear actuators, but found them unsuitable due to insufficient power density and low control frequency. 
After a few iterations, we narrowed our choices to servo motors, ultimately selecting Dynamixel motors for their desirable performance and well-documented support. Given that reproducibility is a hard constraint in our system, we believe a pure Dynamixel design is the most feasible to reproduce, especially for those with limited hardware experience.

For Dynamixel motors, the smallest units start at approximately $50~\mathrm{g}$. This allows us to estimate the total weight as $3100= 30\times50~(\text{motors}) + 600~(\text{computer, battery, camera}) + 1000~(\text{3D-printed structure and metal hardware})~\mathrm{g}$.

To estimate the torque required for each joint to achieve human-like motions, we followed the reference values from \cite{grimmer2020human} and Equation~\ref{eq:performance} to derive the torque estimation:

\begin{equation}
    \bm{\tau}_{\text{robot}} = \frac{\mathrm{h}_{\text{robot}} \cdot \mathrm{m}_{\text{robot}}}{\mathrm{h}_{\text{human}} \cdot \mathrm{m}_{\text{human}}} \cdot \bm{\tau}_{\text{human}}.
\end{equation}

With an estimated height of $0.5~\mathrm{m}$ and weight of $3.1~\mathrm{kg}$, the required lower limb torque for \system to perform the most demanding tasks such as running and slope climbing~\citep{grimmer2020human}, is estimated as follows: ${\tau}_{\text{robot}}^{\text{knee}} = 2.35~\mathrm{Nm},\space  
{\tau}_{\text{robot}}^{\text{ankle\ pitch}} = 2.66~\mathrm{Nm},\space  
{\tau}_{\text{robot}}^{\text{hip\ pitch}} = 1.77~\mathrm{Nm}.$

\begin{table}[ht]
\centering
\caption{Dynamixel Motor Assignments for \system.}
\setlength{\tabcolsep}{0pt}
\resizebox{0.6\linewidth}{!}{
\begin{threeparttable}
\begin{tabular}{@{}lcc@{}} 
\toprule
\textbf{Motor Model} & $\textbf{Stall Torque}^{(a)}$	&\textbf{Assigned DoFs} \\ \midrule
XC330-T288  & 1.0          & Neck PY$^{(b)}$, Waist RY, Hip Y, Gripper \\
XC430-T240BB  & 1.9             & Shoulder P, Ankle R \\
XM430-W210  & 3.0             & Knee P, Ankle P \\ 
2XL430-W250 & 1.5             & Shoulder RY, Elbow RY, Wrist RP \\
2XC430-W250 & 1.8          & Hip RP \\ \bottomrule
\end{tabular}
\begin{tablenotes}
\item[{(a)}] The Stall Torque data are measured at $12\mathrm{V}$ as reported on the Dynamixel official website~\citep{nameintroductiona}. The unit is $\mathrm{Nm}.$
\item[{(b)}] R, P, and Y denote roll, pitch, and yaw respectively.
\end{tablenotes}
\end{threeparttable}
}
\label{tab:dynamixel}
\vspace{0mm}
\end{table}

As shown in Table~\ref{tab:dynamixel}, XM430 is the only option that provides sufficient torque for the knee and ankle pitch joints. Its metal gears, low backdrive resistance, and high torque output make it ideal for these joints that directly impact walking stability.
For the hip, we use 2XC430s for roll and pitch, as they provide sufficient torque while maintaining a compact design, integrating two actuated DoFs in a single housing. This setup allows for a greater range of motion compared to placing two XC430s sequentially.
The XC330 series is the smallest and most cost-effective in the lineup, but has higher tracking errors and backlash. Therefore, we use XC330s on joints with lower torque demands or strict weight and space constraints, such as the neck, waist, hip yaw, and parallel jaw gripper. 
2XL430, a lower-cost variant of the 2XC430, is used in the arm to balance performance and cost, ensuring minimal performance loss while maximizing affordability.
For the remaining joints, including the shoulder pitch and ankle roll, XC430 is used as a standard choice.


\subsection{Power Budget}
\label{sec:power_budget}

\begin{figure}[t]
  \centering
  \includegraphics[width=0.5\linewidth]{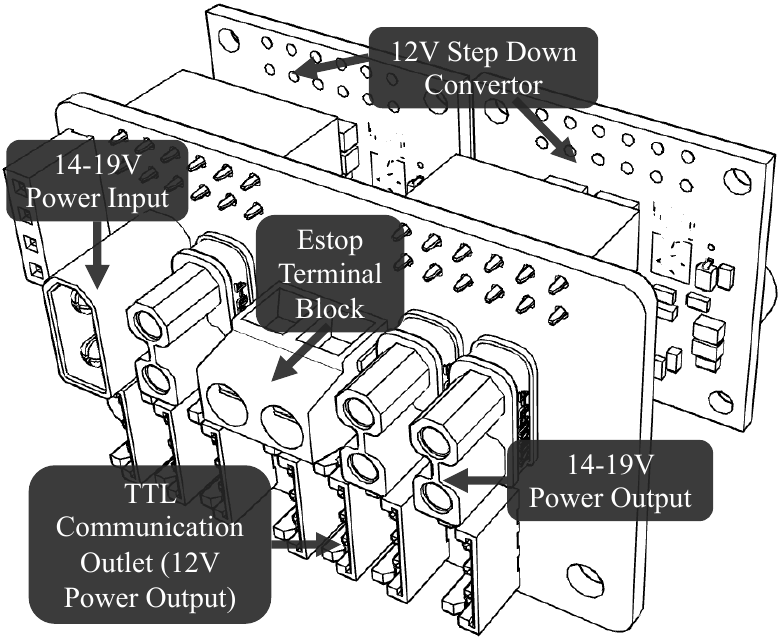}
  \caption{\textbf{Power Distribution.} We show the power distribution board design, including four XT30 power plugs, an Estop terminal block, seven JST EH TTL communication outlets, and two $12\mathrm{V}$ step-down convertors.}
  \label{fig:pcb}
  \vspace{-3mm}
\end{figure}

\begin{figure}[t]
  \centering
  \includegraphics[width=0.6\linewidth]{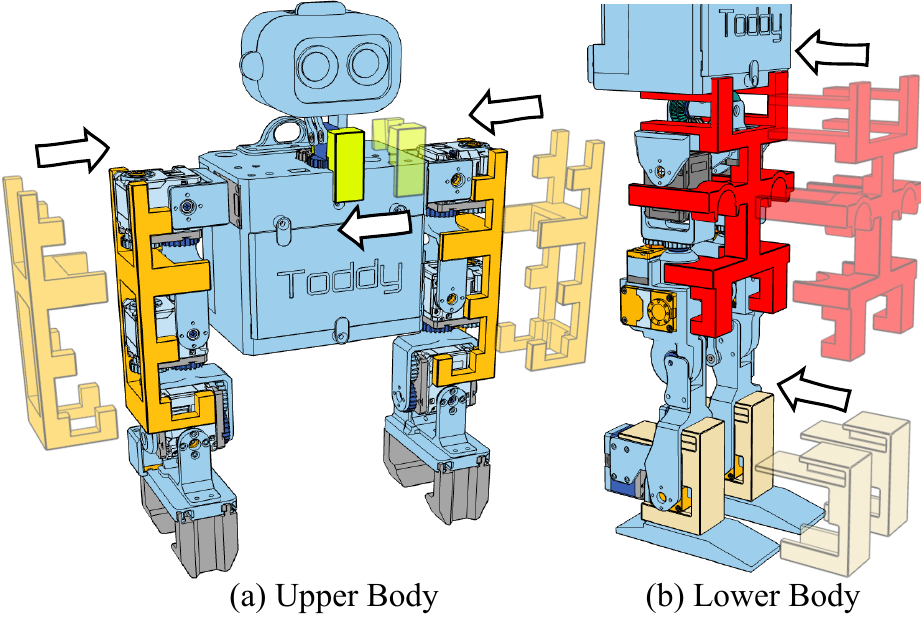}
  \caption{\textbf{Zero-point Calibration.} We 3D-print devices for the plug-and-play zero-point calibration procedure: orange for the arm, yellow for the neck, red for the hip, and beige for the ankle. Arrows indicate the insertion direction, and the zero-point is fixed once the devices click into place.}
  \label{fig:calibration}
  \vspace{-4mm}
\end{figure}

With the motor selection finalized, we can estimate the power budget. The computer and camera together consume approximately $15~\mathrm{W}$ under typical operating conditions. Actual walking power consumption depends on the energy efficiency of the control policy, but preliminary analysis suggests that the upper body requires minimal power during walking, as it carries no additional load. The lower body operates with alternating support, meaning only one leg is actively working at a time. Assuming a 70\% duty cycle, total motor power consumption is estimated at $~50~\mathrm{W}$, requiring a $75~\mathrm{Wh}$ battery for one hour of continuous walking.

For the battery, we offer two options: A $2000~\mathrm{mAh}$ LiPo battery ($215~\mathrm{g}$) available off-the-shelf or a custom-made 4-cell 21700 battery with $5000~\mathrm{mAh}$, which has higher energy density and weighs $330~\mathrm{g}$.
In practice, the battery lasts 3–5 hours in research settings where the robot walks intermittently. The peak power output from the battery is $14.8~\mathrm{V} \times 25~\mathrm{A} = 370~\mathrm{W}$, sufficient to power all the joints. When debugging without a battery, a $15~\mathrm{V}~300~\mathrm{W}$ power supply is a practical alternative—safer and easier to obtain than those required for full-scale humanoids, which often exceed $2~\mathrm{kW}$ and $60~\mathrm{V}$. In practice, \systems battery can last about 2 hours, longer than other commonly available humanoids - Zeroth~\citep{zeroth}: 20 min, OP3~\citep{nameintroduction}:
10-15 min, BRUCE~\citep{liu2022design}: 20 min, Berkeley Humanoid Lite~\citep{chi2025demonstrating}: 30 min, NAO H25~\citep{nao}: 1 hour, Unitree G1~\citep{unitreea}: 2 hours.

As shown in Figure~\ref{fig:pcb}, the battery provides a $14-19\mathrm{V}$ input, regulated to $12\mathrm{V}$ via dual step-down converters to power the motors through TTL communication outlets. An E-stop terminal block controls motor power, enabling emergency reboots. The $14-19\mathrm{V}$ output powers the Jetson Orin NX, which remains on when the battery is connected to prevent data loss from abrupt shutdown.


\subsection{Zero-point Calibration}
\label{sec:calibration}

Figure~\ref{fig:calibration} shows the zero-point calibration process with the 3D-printed devices.

\subsection{Motor Test Bed}
\label{sec:motor_test_bed}

Each motor family used on \system differs in gearbox material, gear ratio, and core, resulting in distinct stiction, backdrive resistance, damping, and inertia. To enable successful sim-to-real transfer, we must minimize the gap between simulated and real actuator behavior.

Specifically, MuJoCo/MJX models a simplified actuator characteristics with predominantly 3 values: frictionloss, damping, and armature~\citep{todorov2012mujoco}. Frictionloss is the minimum torque $\bm{\tau_{f}}$ required for the actuator to start moving, with a unit of $\mathrm{Nm}$. Damping $\bm{\mathrm{d}}$ refers to the rate at which backdrive resistance increases with speed, measured in $\mathrm{Nms/rad}$. Armature $I$ denotes the effective rotor inertia, accounting for the gearbox, with units of $\mathrm{kgm^2}$. The model in MuJoCo does not consider stiction peak, therefore, the actuator resistance is simply:
\begin{equation}
    \bm{\tau_r} = \bm{\tau_{f}} + \bm{\mathrm{d}}\cdot\bm{\dot{\mathrm{q}}}
\end{equation}

Therefore, it is straightforward to design a test platform to measure these physical parameters using a series of test sequences. Specifically, damping and friction loss can be measured by backdriving the motor at a constant RPM and recording the resisting torque via a torque sensor. A linear fit to the torque-speed data yields friction loss as the intercept at 0 RPM and damping as the slope. To estimate armature inertia, the actuator is allowed to spin freely before cutting motor power to observe spin-down behavior. Using the previously measured damping values, the resistance power can be numerically integrated to estimate the initial stored energy $E$. The armature inertia $I$ is then computed using:

\begin{equation}
    E = \frac{1}{2}I\cdot \omega^2 \rightarrow I = \frac{2E}{\omega^2}
\end{equation}

\begin{figure}
  \centering
  \includegraphics[width=0.6\linewidth]{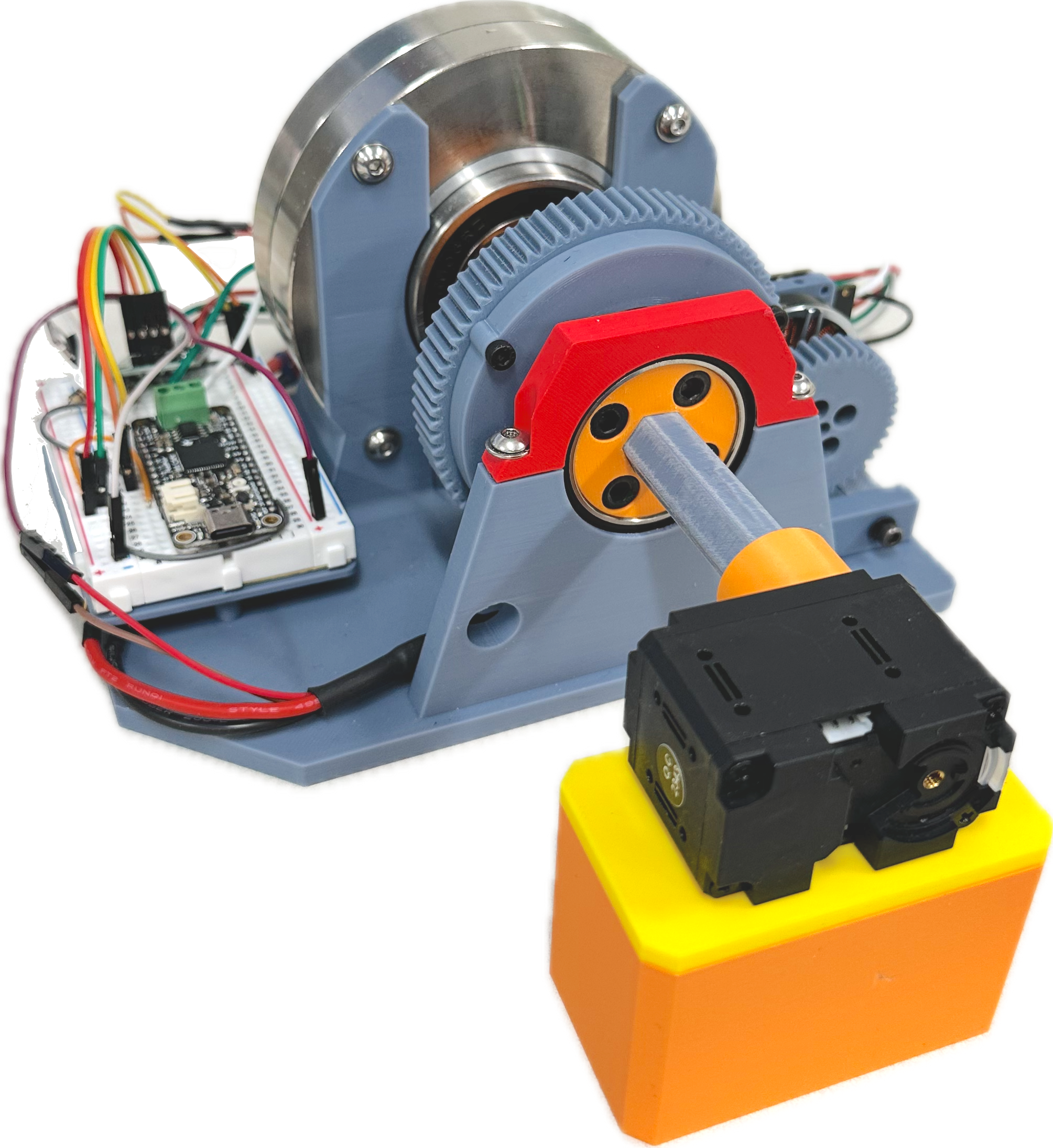}
  \caption{\textbf{Motor Test Stand.} We show a photo of the assembled motor test stand. The test motor is mounted via the blue-grey quick-connect shaft, which is directly coupled to a torque sensor. A secondary motor on the side provides active driving torque, while a powder brake at the rear offers up to $5~\mathrm{Nm}$ of controllable resistance. On the left, a controller MCU handles torque sensing, brake actuation, and CAN communication with the driving motor.}
  \label{fig:mts}
  \vspace{-2mm}
\end{figure}

We have developed the test profile and analysis code to automatically identify these parameters in MuJoCo. If a more complex actuator model is needed, our modular test bed can be easily extended with additional sensors. The current setup supports active braking ($5~\mathrm{Nm}$), active driving ($1~\mathrm{Nm}$), and accurate torque measurements (precision $3e^{-4}~\mathrm{Nm}$), meeting most motor SysID requirements. The test bed will be open-sourced with the release. A photo of the setup is shown in Figure~\ref{fig:mts}.

\subsection{Actuation Model}
\label{sec:actuation_model}

The PD position control equation is computed through:
\begin{equation}
\bm{\tau}_m = \bm{{\mathrm{k}}}_p (\bm{{\hat{\mathrm{q}}}}-\bm{\mathrm{q}}) - (\bm{{\mathrm{k}}}_{d}^{\min}+\bm{{\mathrm{k}}}_d) \bm{\dot{\mathrm{q}}},
\end{equation}
where $\bm{{\mathrm{k}}}_p$ and $\bm{{\mathrm{k}}}_d$ are the gains, $\bm{\hat{\mathrm{q}}}$ is the joint setpoint, $\bm{{\mathrm{q}}}$ is the joint position, and $\bm{\dot{\mathrm{q}}}$ is the joint velocity.

During motor testing, we observed significant additional damping when the motor was powered on, even with $\mathrm{k}_d = 0$. We modeled this effect as $\mathrm{k}_{d}^{\min}$, separate from passive damping, as it only occurs when the motor is active and should respect torque limits. Additionally, we found the conversion factor from Dynamixel’s unitless $k_p$ to physical $k_p$ used in simulators to be approximately 150.

\begin{figure}
  \centering
  \includegraphics[width=0.7\linewidth]{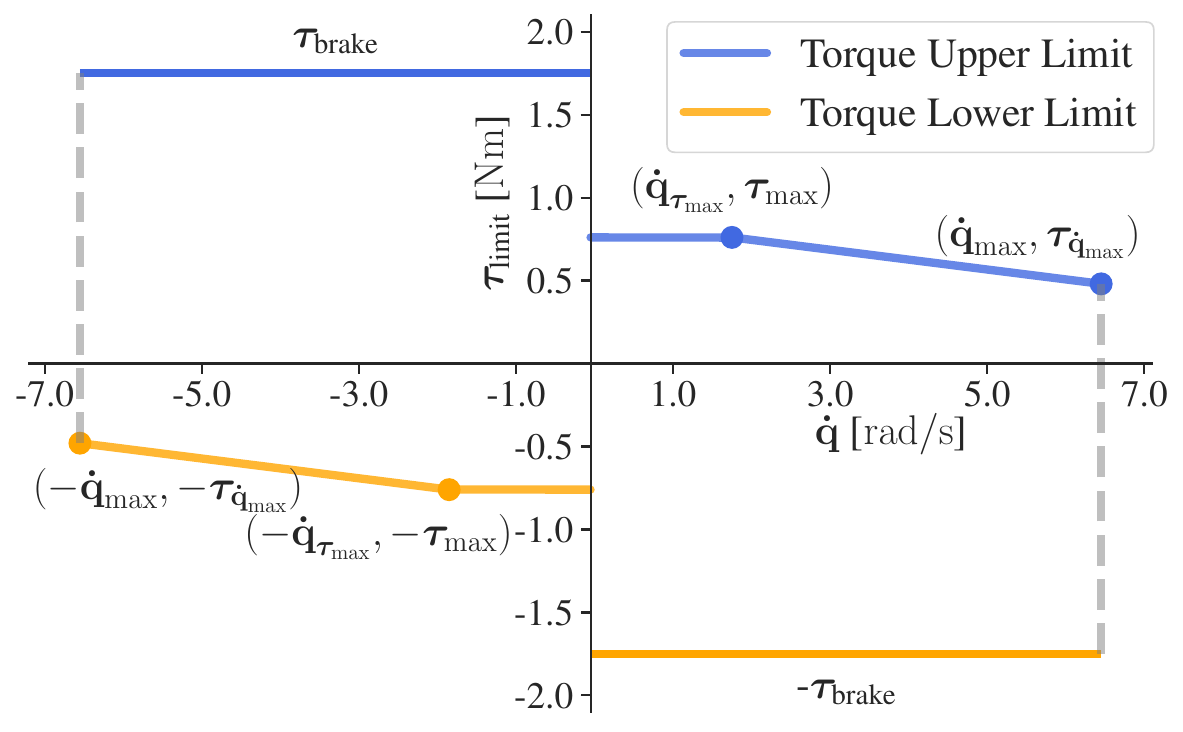}
  \caption{The relationship between torque limit ${\tau}_{\text{limit}}$ and joint velocity ${\dot{\mathrm{q}}}$ for Dynamixel XC330.}
  \label{fig:tau_qdot}
  \vspace{-3mm}
\end{figure}

Inspired by \citet{grandia2024design}, we design our actuation model as follows. The motor torque limit $\bm{\tau}_{\text{limit}}$ varies with velocity: 

\begin{equation}
\bm{\tau}_{\text{limit}} = 
\begin{cases} 
\bm{\tau}_{\max}, & |\bm{\dot{\mathrm{q}}}| \leq \bm{\dot{\mathrm{q}}}_{\bm{\tau}_{\max}} \\
\cfrac{\bm{\dot{\mathrm{q}}}_{\max} - |\bm{\dot{\mathrm{q}}}|}{\bm{\dot{\mathrm{q}}}_{\max} - \bm{\dot{\mathrm{q}}}_{\bm{\tau}_{\max}}} \cdot \bm{\tau}_{\max}, & \bm{\dot{\mathrm{q}}}_{\bm{\tau}_{\max}} < |\bm{\dot{\mathrm{q}}}| \leq \bm{\dot{\mathrm{q}}}_{\max} \\
0, & |\bm{\dot{\mathrm{q}}}| > \bm{\dot{\mathrm{q}}}_{\max}
\end{cases}
\end{equation}
$\bm{\tau}_{\text{limit}}$ comprises a constant torque limit $\bm{\tau}_{\max} > 0$ for low velocities, and a linear reduction in available torque beyond a specific velocity $\bm{\dot{\mathrm{q}}}_{\bm{\tau}_{\max}}$. This linear limit reaches zero torque at the velocity $\bm{\dot{\mathrm{q}}}_{\max}$.
Note that $\bm{\tau}_{\text{limit}}$ is the maximum acceleration torque, but maximum deceleration torque is assumed to be always a constant $\bm{\tau}_{\text{brake}}$. We separate the braking torque limit rather than using the acceleration torque limit as in \citet{grandia2024design}, since the motor typically provides higher braking torque due to passive resistance and gearbox inefficiencies. Additionally, the gearbox efficiency has a large effect when the motor is back-driven by external torque. Concretely, when the motor is back-driven, it shows significantly higher $k_p$, as the external torque is consumed by the gearbox inefficiency. We model this by a parameter called passive-active ratio, which is equivalent to $\frac{1}{\eta^2}$, where $\eta$ is gearbox efficiency. We have empirically determined the passive-active ratio to be 3, equating gearbox torque efficiency around 58\%. The resulting sysID relationship for Dynamixel XC330 is illustrated in Figure~\ref{fig:tau_qdot}. 


\begin{table}[t]
\centering
\caption{SysIDed Parameters for Dynamixel Motors.}
\setlength{\tabcolsep}{6pt}
\resizebox{0.75\linewidth}{!}{
\begin{threeparttable}
\begin{tabular}{lccccc}
\toprule
\textbf{Parameter}       & \textbf{2XL430} & \textbf{XC330} & \textbf{XC430} & \textbf{2XC430} & \textbf{XM430-W210} \\ 
\midrule
Damping$^{(a)}$                 & 0.0010        & 0.0036       & 0.0066       & 0.0028        & 0.0056       \\ 
Armature                        & 0.0083        & 0.0040       & 0.0042       & 0.0044        & 0.0022       \\ 
Friction Loss                   & 0.078         & 0.036        & 0.024        & 0.060         & 0.025        \\ 
${\tau}_{\max}$                & 0.94          & 0.76         & 1.32          & 1.09           & 1.61          \\ 
${\dot{\mathrm{q}}}_{{\tau}_{\max}}$  & 2.00           & 1.80          & 1.60          & 2.00           & 0.10         \\ 
${\dot{\mathrm{q}}}_{\max}$         & 5.97           & 6.50          & 7.00          & 6.78           & 7.63          \\ 
$\tau_{\dot{\mathrm{q}}_{\max}}$         & 0.10          & 0.48         & 0.21         & 0.23          & 0.47         \\ 
$\mathrm{k}_d^{\text{min}}$              & 0.161          & 0.384         & 0.170         & 0.185          & 0.203         \\ 
$\tau_{\text{brake}}$           & 1.40           & 1.75          & 3.00          & 2.20           & 3.70          \\ 
\bottomrule
\end{tabular}

\begin{tablenotes}
\item[{(a)}] The units of damping and $\mathrm{k}_d^{\text{min}}$ are $\mathrm{Nms/rad}$, the unit of armature is $\mathrm{kgm^2}$, the units of frictionloss, ${\tau}_{\max}$ and $\tau_{\text{brake}}$ are $\mathrm{Nm}$, and the units of ${\dot{\mathrm{q}}}_{{\tau}_{\max}}$ and ${\dot{\mathrm{q}}}_{\max}$ are $\mathrm{rad/s}$.
\end{tablenotes}

\end{threeparttable}
}
\label{tab:motor_parameters}
\vspace{-3mm}
\end{table}

The joint torque is calculated by applying torque limits to $\bm{\tau}_m$ and combining the resistance force $\bm{\tau}_r$:

\begin{equation}
\bm{\tau} = 
\begin{cases} 
\text{clamp}_{[-\bm{\tau}_{\text{max}}, \bm{\tau}_{\text{brake}}]}(\bm{\tau}_m) - \bm{\tau}_r,
& \bm{\dot{\mathrm{q}}} \geq 0 \\
\text{clamp}_{[-\bm{\tau}_{\text{brake}}, \bm{\tau}_{\text{max}}]}(\bm{\tau}_m) + \bm{\tau}_r,
& \bm{\dot{\mathrm{q}}} < 0
\end{cases}
\end{equation}

$\bm{\tau}_r$ follows the joint passive force model in MuJoCo~\citep{todorov2012mujoco}, which is characterized by three parameters: damping, armature, and friction loss.
The sysIDed parameters for various Dynamixel motors are presented in Table~\ref{tab:motor_parameters}.

Using this actuator model, we jointly optimize all parameters to minimize the gap between simulated and real-world tracking. The model includes 9 parameters: damping, frictionloss, armature, $\bm{\tau}_{\max}$, 
$\bm{\dot{\mathrm{q}}}_{\bm{\tau}_{\max}}$, 
$\bm{\tau}_{\bm{\dot{\mathrm{q}}}_{\max}}$, 
$\bm{\dot{\mathrm{q}}}_{\max}$, 
$\bm{\mathrm{k}}_{d}^{\min}$, and 
$\bm{\tau}_{\text{brake}}$. Due to the complexity of this optimization, constraining parameter ranges is crucial for convergence to a valid local minimum. The final simulation achieves an average tracking error of 1.3° with the optimized parameters (Table~\ref{tab:motor_parameters}).
\subsection{Pupperteering Mapping}
\label{sec:joystick}

\begin{figure}
  \centering
  \includegraphics[width=0.8\linewidth]{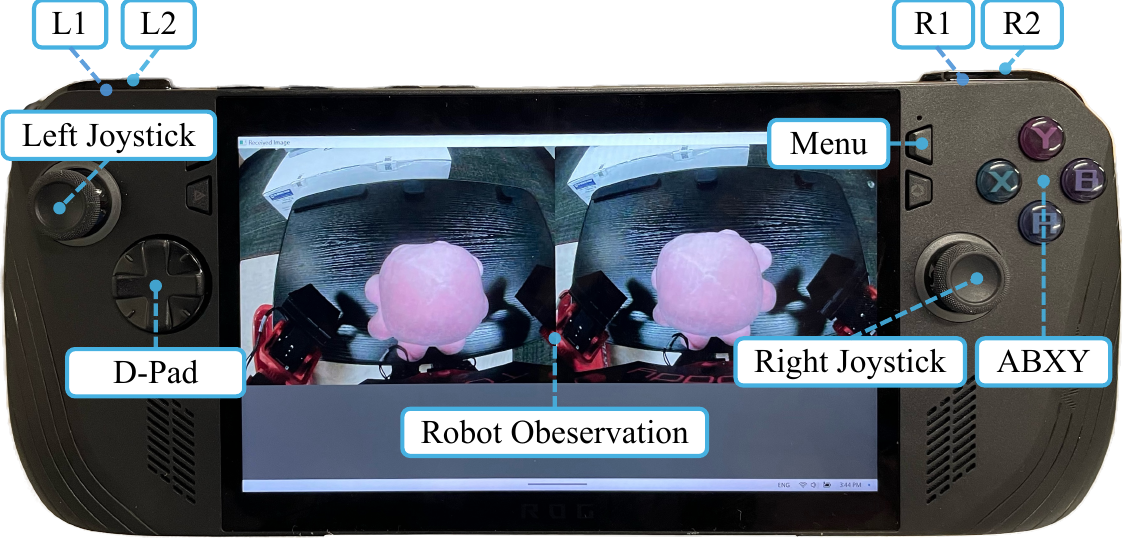}
  \caption{\textbf{Remote Controller Layout}. We show the button and axis layout on ROG Ally X.}
  \label{fig:joystick}
  \vspace{-3mm}
\end{figure}
\begin{table}[t]
\centering
\caption{Pupperteering Button and Axis Mapping}
\setlength{\tabcolsep}{3pt}
\resizebox{0.7\linewidth}{!}{
\begin{threeparttable}
\begin{tabular}{@{}lp{6.5cm}@{}} 
\toprule
\textbf{Button or Axis$^{(a)}$} & \textbf{Effect} \\ \midrule
Menu  & Toggle teleoperation / Mark episode start or end during data collection \\
Left Joystick $\updownarrow$ & Walk forward or backward along the x-axis \\
Left Joystick $\leftrightarrow$  & Walk leftward or rightward along the y-axis \\ 
Right Joystick $\updownarrow$ & Stand up or squat down \\
Right Joystick $\leftrightarrow$  & Turn clockwise or counterclockwise around the z-axis \\
D-Pad $\updownarrow$ & Lean with the waist roll joint \\
D-Pad $\leftrightarrow$ & Twist with the waist yaw joint \\
Y and A & Look up or down with the neck pitch joint \\
X and B & Look left or right with the neck yaw joint \\
L1 & Hold to run the bimanual DP and release to end. \\
R1 & Hold to run the full-body DP and run to end. \\
L2 & Hold to run the wagon pushing policy and release to end. \\
R2 & Hold to run the cuddling policy and release to end. \\
\bottomrule
\end{tabular}
\begin{tablenotes}
\item[{(a)}] This mapping is compatible with various remote controllers and has been tested on the ROG Ally X and Steam Deck. The remaining buttons can be assigned to additional skills based on user preference.
\end{tablenotes}

\end{threeparttable}
}
\label{tab:mapping}
\vspace{-2mm}
\end{table}

Figure~\ref{fig:joystick} illustrates the remote controller layout to teleoperate \system, with button and axis mappings detailed in Table~\ref{tab:mapping}. During teleoperation, the human operator sends velocity commands to the walking policy and determines the timing for skill transitions. The same mapping was used for Steam Deck and ROG Ally X.

\subsection{Reinforcement Learning Details}
\label{sec:rl_details}

\begin{table}[t]
\centering
\caption{Hyperparameters for PPO Training.}
\setlength{\tabcolsep}{7pt}
\begin{threeparttable}
\begin{tabular}{lc}
\toprule
\textbf{Parameter} & \textbf{Value} \\ 
\midrule
Policy hidden layer sizes & $(512, 256, 128)$ \\ 
Value hidden layer sizes & $(512, 256, 128)$ \\ 
Number of timesteps & $3\times10^8$ \\ 
Number of environments & $1024$ \\ 
Episode length & $1000$ \\ 
Unroll length & $20$ \\ 
Batch size & $256$ \\ 
Number of minibatches & $4$ \\ 
Number of updates per batch & $4$ \\ 
Discounting factor & $0.97$ \\ 
Learning rate & $0.0001$ \\ 
Entropy cost & $0.0005$ \\ 
Clipping epsilon & $0.2$ \\ 
\bottomrule
\end{tabular}
\end{threeparttable}
\label{tab:hyperparameters}
\vspace{-2mm}
\end{table}
\begin{table}[t]
\centering
\caption{Reward Shaping for PPO Training.}
\setlength{\tabcolsep}{7pt}
\begin{threeparttable}
\begin{tabular}{lc}
\toprule
\textbf{Imitation Term} & \textbf{Value} \\ 
\midrule
Torso quaternion & $1.0$ \\ 
Linear velocity (XY) & $5.0$ \\ 
Linear velocity (Z) & $1.0$ \\ 
Angular velocity (XY) & $2.0$ \\ 
Angular velocity (Z) & $5.0$ \\ 
Leg motor position & $5.0$ \\
Feet contact & $1.0$ \\ 
\midrule
\textbf{Regularization Term} & \\ 
\midrule
Feet air time & $500.0$ \\
Feet clearance & $0.05$ \\
Feet distance & $1.0$ \\
Feet slip & $0.05$ \\
Align with the ground & $1.0$ \\
Stand still & $1.0$ \\
Torso roll & $0.5$ \\ 
Torso pitch & $0.5$ \\
Collision & $0.1$ \\
Leg action rate & $0.05$ \\ 
Leg action acceleration & $0.05$ \\ 
Motor torque & $0.01$ \\ 
Energy & $0.05$ \\ 
\midrule
\textbf{Survival Term} & \\ 
\midrule
Survival & $10.0$ \\ 
\bottomrule
\end{tabular}
\end{threeparttable}

\label{tab:reward_scales}
\vspace{-1mm}
\end{table}

The RL implementation leverages MJX~\citep{todorov2012mujoco} and Brax~\citep{brax2021github}. We train the policy using PPO~\citep{schulman2017proximal} with hyperparameters listed in Table~\ref{tab:hyperparameters}.
Inspired by prior work~\citep{gu2024humanoidgym, gu2024advancing}, our reward function is shaped by three categories of reward terms as detailed in Table~\ref{tab:reward_scales}. 
Full implementation details are available in our open-source codebase.
During inference, the RL policy runs on the CPU of Jetson Orin NX 16GB, achieving a $50~\mathrm{Hz}$ control loop.
\subsection{Diffusion Policy Details}
\label{sec:dp_details}


The diffusion policy processes a cropped and downsampled $96\times96$ RGB image, which is encoded by a ResNet~\citep{he2016deep} pretrained on ImageNet~\citep{deng2009imagenet} to extract visual features. Both leader and follower joint angles are downsampled to $10~\mathrm{Hz}$ for training, where the leader joint angles serve as actions and the follower as observations.
To prevent motor overload during data collection, the upper body motors use low proportional gains, allowing modulation of the manipulation force. This behavior is embedded in the discrepancy between leader and follower joint angles, which the policy learns.



The model is trained with 100 diffusion steps. During inference, the trained model runs directly on the Jetson Orin NX 16GB with 3 DDPM steps, which are sufficient for satisfactory results. With 300M parameters, the inference latency remains under $0.1~\mathrm{s}$ on the GPU, ensuring smooth execution at $10~\mathrm{Hz}$ without stuttering. Each inference yields a 16-step prediction; the first 3 actions are discarded to compensate for latency~\citep{chi2024universal}, and the next 5 actions are executed.

\subsection{Velocity Tracking Details}
\label{sec:vel_tracking}

Table~\ref{tab:tracking_errors} shows the quantitative results of the walking velocity tracking experiment.

\begin{table}[ht]
\centering
\caption{Tracking Errors in Simulation versus the Real World}
\resizebox{0.5\linewidth}{!}{
\begin{tabular}{@{}lcc@{}}
\toprule
    Tracking Errors        & Simulation & Real-World \\ \midrule
    Position [m]                  & 0.082      & 0.133 $\pm$ 0.018 \\
    Linear Velocity [m/s]           & 0.016      & 0.032 $\pm$ 0.002 \\
    Angular Velocity [rad/s]          & 0.056      & 0.113 $\pm$ 0.010 \\ \bottomrule
\end{tabular}
}
\vspace{-1mm}
\label{tab:tracking_errors}
\end{table}

\subsection{Skill Chaining Details}
\label{sec:skill_chaining}

\begin{figure*}[h]
  \centering
  \includegraphics[width=\linewidth]{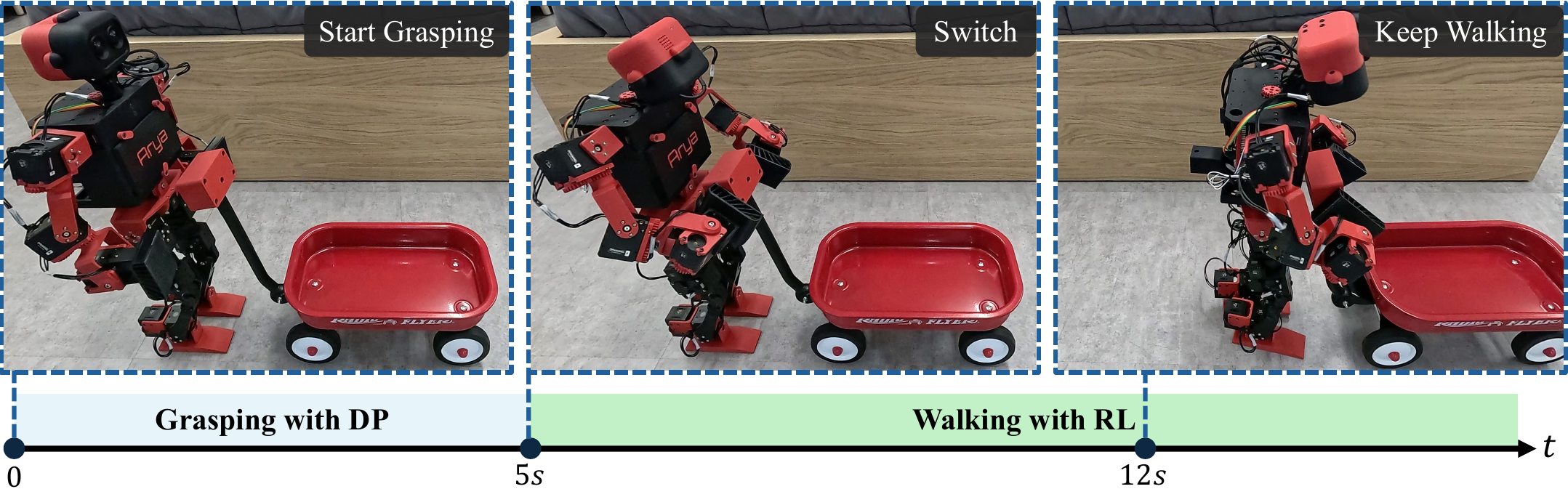}
  \caption{\textbf{Skill Chaining.} We demonstrate that \system can seamlessly transition from a DP-based grasping skill to an RL-trained walking policy, enabling it to grasp the wagon handle and push it forward while maintaining its grip.}
  \label{fig:skill_chaining}
  \vspace{-5mm}
\end{figure*}

Figure~\ref{fig:skill_chaining} shows the results of the skill chaining experiment.

\subsection{Reproduction Results}
\label{sec:reproduction}

\begin{figure}[h]
  \centering
  \includegraphics[width=\linewidth]{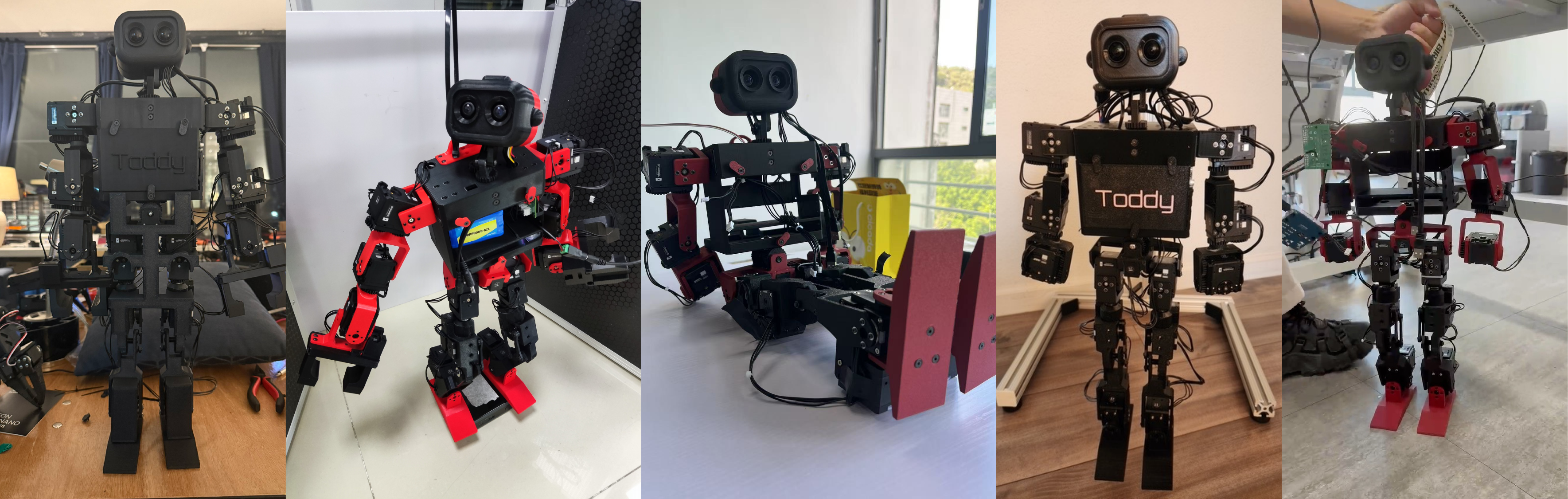}
  \caption{\textbf{Successful Reproductions.} We showcase five independent replications of \system by teams across the globe. On average, each team completed the replication in about a week.}
  \label{fig:reproduction}
  \vspace{-1mm}
\end{figure}

Figure~\ref{fig:reproduction} shows some successful reproductions of \system by teams around the world.

\begin{figure*}
  \centering
  \includegraphics[width=\linewidth]{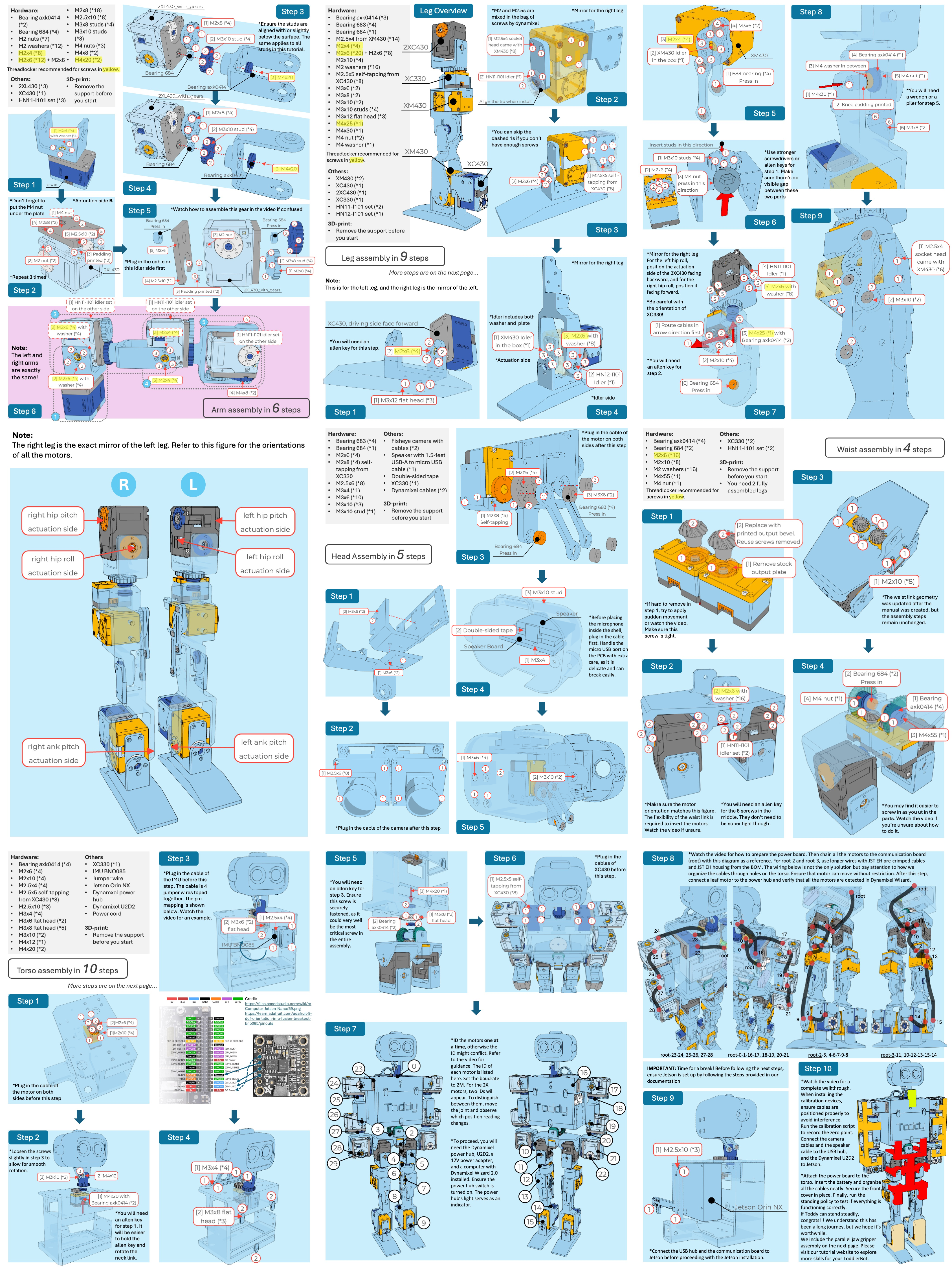}
  \caption{\textbf{Assembly Manual.} We present the open-source assembly manual for \system.}
  \label{fig:manual}
  \vspace{0mm}
\end{figure*}

Figure~\ref{fig:manual} shows our open-source assembly manual.

\end{document}